\pdfoutput=1
\documentclass{article} %
\usepackage{iclr2027_conference,times}

\usepackage[utf8]{inputenc} %
\usepackage[T1]{fontenc}    %
\usepackage{hyperref}       %
\usepackage{url}            %
\usepackage{booktabs}       %
\usepackage{amsfonts}       %
\usepackage{nicefrac}       %
\usepackage{microtype}      %
\usepackage{xcolor}         %
\usepackage{amsmath}
\usepackage{multirow}
\usepackage{graphicx}
\usepackage{algorithm, algpseudocode}
\usepackage{wrapfig}
\usepackage{subcaption}
\usepackage{enumitem}
\usepackage[capitalize,noabbrev]{cleveref}
\usepackage{lipsum}
\setcitestyle{maxnames=2}

\usepackage{graphicx}
\usepackage{xcolor}

\definecolor{baseblue}{HTML}{0066CC}
\definecolor{mergeyellow}{HTML}{D4B000}
\definecolor{mergeorange}{HTML}{F07800}
\definecolor{mergered}{HTML}{D7191C}

  \usepackage{tcolorbox}

\title{zip2zip++: Towards Inference-Time Token Compression with Minimal Quality Loss}

\author{%
  Saibo Geng$^{1}$\thanks{Equal contribution. Author order was determined by a dice roll.}
  \quad \textbf{Andrea Gentilini}$^{2,3}$\footnotemark[1]\hspace{0.35em}\textsuperscript{,}\thanks{Work done during an internship at EPFL. Andrea Gentilini is supported by the Konrad Zuse School of Excellence in Learning and Intelligent Systems (ELIZA) through the DAAD programme Konrad Zuse Schools of Excellence in Artificial Intelligence, sponsored by the Federal Ministry of Education and Research.}
  \quad \textbf{Xinxian Ma}$^{1}$\footnotemark[1] 
  \quad \textbf{Nathan Ranchin}$^{4}$\\
  \textbf{Yunzhen Yao}$^{1}$
  \quad \textbf{Robert West}$^{1}$ \\
  $^{1}$EPFL \quad
  $^{2}$Technical University of Munich \quad
  $^{3}$Zuse School ELIZA \quad
  $^{4}$ETH Z\"urich \\
  \texttt{\{saibo.geng, xinxian.ma, yunzhen.yao, robert.west\}@epfl.ch} \\
  \texttt{a.gentilini@tum.de} \quad
  \texttt{nathan.ranchin@ethz.ch}
}

\iclrfinalcopy %
\begin{document}

\maketitle
\lhead{Preprint}   %

\begin{abstract}
Dynamic vocabularies built on-the-fly via Lempel--Ziv--Welch (LZW) compression let a
language model represent the same text in fewer tokens, but prior work shows that
this efficiency comes at a cost to reasoning and perplexity. We show that this quality gap is not a fundamental price of compression but can be substantially closed through targeted improvements to the compression layer and LZW codec, without modifying the underlying pretrained language model.
Our analysis reveals four key issues: (1)~LZW's compression algorithm 
creates an implicit conflict between the input embedding and the output embedding — a tension that a single shared embedding cannot resolve; (2)~naively applying RoPE to LZW-compressed sequences ignores that merged tokens can span multiple tokens; (3) merging certain tokens (notably digits) disproportionately hurts downstream accuracy; and (4)~dynamic vocabularies introduce an
inherent segmentation ambiguity that confounds standard evaluation.
We address these through decoupled input/output dynamic embedding layer, size-aware RoPE positioning, and a
modified LZW codec supporting selective merges and multi-view segmentations.
Across four models, our recipe improves perplexity on every corpus we evaluated as well as mathematical reasoning accuracy, closing roughly half the quality gap between zip2zip and the uncompressed baseline at a similar compression ratio on general text. This represents a significant step toward lossless adaptive-vocabulary compression for language models. Code and models are available at \url{https://github.com/epfl-dlab/zip2zip-plus-plus}.
\end{abstract}

\section{Introduction}

The vocabulary of a language model determines the granularity at which text is represented, directly governing both computational efficiency and representational capacity. A more compact tokenization—one that represents the same text in fewer tokens—yields immediate practical benefits: lower inference cost (and, by extension, lower end-user billing in API-based services), reduced KV-cache memory footprint, and a larger effective context window within the same fixed token budget. Modern large language models universally adopt fixed vocabularies derived from statistical compression algorithms—most commonly Byte Pair Encoding (BPE) \citep{sennrich-etal-2016-neural} or its variants \citep{kudo-richardson-2018-sentencepiece,kumar-thawani-2022-bpe,liu-etal-2025-superbpe}—trained once on large web corpora and frozen thereafter. While this approach is simple and effective on average, it is inherently suboptimal for any particular input: domain-specific terminology, low-resource languages, and structured data (e.g., code, mathematical expressions) are often over-segmented, producing unnecessarily long sequences and wasting compute \citep{rust-etal-2021-good, ahia-etal-2023-languages,foroutan-etal-2026-parity}. Every excess token directly translates to unnecessary autoregressive steps, increased latency, and wasted context capacity. Prior attempts to mitigate this using vocabulary expansion \citep{wang-etal-2019-improving}, domain-adaptive retraining \citep{adapt-llama-chinese}, or efficient language-specific vocabulary expansion \citep{kim2024efficienteffectivevocabularyexpansion_korean}—require per-domain intervention and do not generalize to new domains.

Dynamic vocabularies, which adapt tokenization to the input at inference time offer a principled alternative. Rather than committing to a single fixed segmentation, such methods construct input-specific tokens on the fly, achieving shorter sequences without retraining. One line of work removes the fixed 
vocabulary entirely by operating at the byte level~\citep{yu2023megabyte,slagle2024spacebyte,blt, hnet,feher-etal-2025-retrofitting}, learning to dynamically group bytes into variable-length patches through the model architecture 
itself. Another line of work~\citep{liu-etal-2024-generation, geng2025zip2zip} builds on top of the existing BPE token vocabulary, constructing higher-order tokens on the fly from observed token sequences. \citet{geng2025zip2zip} recently repurposed the Lempel--Ziv--Welch (LZW) streaming compression codec~\citep{welch1984technique} for language model inference, dynamically building a \textbf{hypertoken} vocabulary on the fly. Unlike BPE, which learns a fixed merge table offline from a static corpus, LZW expands its dictionary online from the input token stream, reducing token usage by up to 40\%.
However, a language model's vocabulary is tightly coupled with its architecture through the embedding and output projection layers, so any vocabulary change risks degrading model behavior. \citet{geng2025zip2zip} mitigate this via lightweight LoRA finetuning~\citep{hu2022lora}, but at a significant cost: perplexity increases substantially and downstream performance degrades sharply on some tasks.

In this work, we identify the key sources of quality loss in \texttt{zip2zip} and address them with four targeted modifications:
(1)~\textbf{Decoupled Residual Hypertoken Embeddings} with identity initialization (\Cref{sec:untied,sec:residual});
(2)~\textbf{Size-Aware RoPE}, which positions a hypertoken by the span of base tokens it covers (\Cref{sec:rope});
(3)~\textbf{Fault-Tolerant LZW}, a codec that stays robust when the compressor is an imperfect neural model and lets fragile tokens be exempted from merging (\Cref{sec:lzw});
and (4)~\textbf{Multi-View Perplexity}, which marginalizes over every segmentation that decodes to the target rather than only the canonical one (\Cref{sec:fault-tolerant}).
Together, these substantially narrow the gap to uncompressed baselines.

\section{Background: Zip2Zip}

\paragraph{Tokenizer: LZW-based dynamic vocabulary.}
Rather than relying solely on a fixed base vocabulary $\mathcal{V}$ of size $|\mathcal{V}|$,
\texttt{zip2zip} augments it with a dynamic \emph{hyper-vocabulary} $\mathcal{V}_h$
constructed online via the Lempel--Ziv--Welch (LZW) compression algorithm.
Given an input sequence of base tokens $\mathbf{x} = (x_1, x_2, \dots, x_T)$ where
each $x_i \in \mathcal{V}$, LZW incrementally identifies recurring subsequences
$x_{1:K} := x_1 \dots x_K$ with $K \leq M$, and merges them into
\emph{hypertokens} $z \in \mathcal{V}_h$, where $M$ is the maximum merge size controlling the
granularity of compression.
We refer to $x_1, \dots, x_K$ as the \emph{constituent} base tokens of $z$, and
to $K$ as its \emph{span length}.
LZW operates in a streaming fashion: a hypertoken created at step $t$
is immediately available for reuse at step $t+1$. The resulting compressed
sequence $\mathbf{z} = (z_1, z_2, \dots, z_{T'})$, where each $z_i \in
\mathcal{V} \cup \mathcal{V}_h$, serves as the model's input and
prediction target. At any point during decoding, the model operates over the
combined vocabulary $\mathcal{V} \cup \mathcal{V}_h$, where $\mathcal{V}_h$ grows
dynamically as new patterns are encountered.

\paragraph{Architecture: Dynamic Embedding.}
Since hypertokens are not part of the original vocabulary $\mathcal{V}$,
they lack entries in the static embedding matrix
$\mathbf{W}_{\text{in}} \in \mathbb{R}^{|\mathcal{V}| \times d}$ and
the unembedding matrix
$\mathbf{W}_{\text{out}} \in \mathbb{R}^{|\mathcal{V}| \times d}$.
\texttt{zip2zip} introduces a lightweight \emph{hyper-encoder}---a
small transformer $f_\phi$---that computes hypertoken embeddings
on-the-fly from their constituent base token embeddings:
\begin{equation}
  v(z) = f_\phi\bigl(
    \mathbf{W}[x_1],\;
    \mathbf{W}[x_2],\;
    \dots,\;
    \mathbf{W}[x_K]
  \bigr),
\end{equation}
where $v(z) \in \mathbb{R}^d$ is fed into the model in the same way as the base token embeddings. The output side is
handled symmetrically: a hyper-encoder (optionally tied with the input
one) maps rows of $\mathbf{W}_{\text{out}}$ to an unembedding vector
$v_{\text{out}}(z)$, enabling the model to produce logits over the
full union $\mathcal{V} \cup \mathcal{V}_h$ at each decoding step: the
logit for emitting $z$ at position $j$ is $\mathbf{h}_j^\top v_{\text{out}}(z)$,
where $\mathbf{h}_j \in \mathbb{R}^d$ is the decoder's hidden state at
that position.

\paragraph{Training: Efficient Uptraining.}
An existing pretrained LLM can be uptrained to \texttt{zip2zip} via
continued pretraining on LZW-compressed sequences using the standard
causal language modeling objective over the compressed token space
$\mathcal{V} \cup \mathcal{V}_h$. With parameter-efficient finetuning
(LoRA;~\citealp{hu2022lora}), only the hyper-encoder $f_\phi$ and
low-rank adapters $\Delta\theta$ are trained, requiring as few as
100M tokens.

\section{Hypertoken Encoder Architecture}

\subsection{Prefix-Aligned Geometry of Hypertoken Embeddings}
\label{sec:untied}

\begin{figure}[h]
\centering
\includegraphics[width=\textwidth]{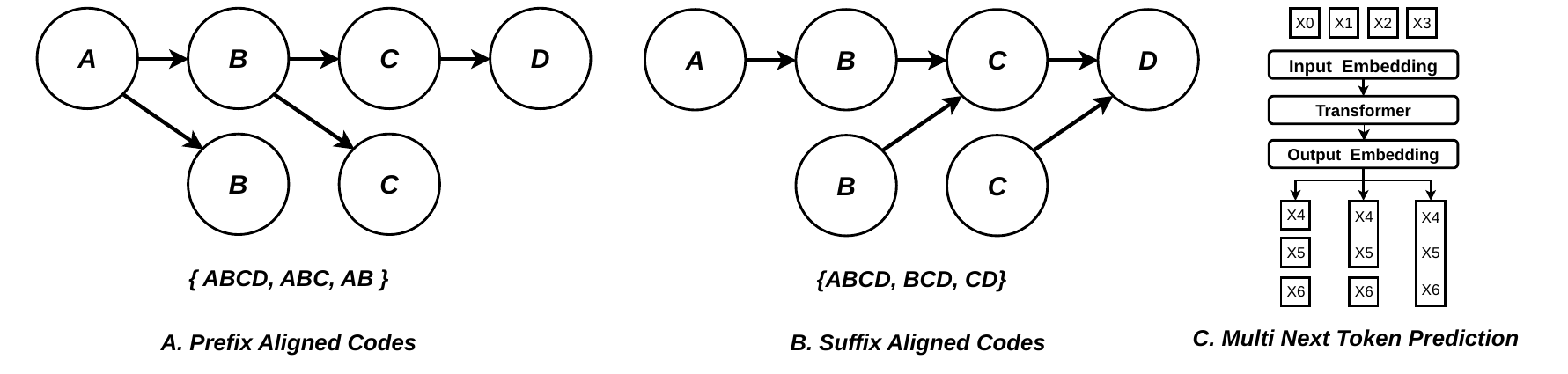}
\caption{\textbf{Why LZW induces prefix-aligned output codes.}
\textbf{(a)}~The LZW dictionary is prefix-closed: an entry is registered only
after its own prefix, so the entries available at a given position form a chain
under the prefix order ($\{AB, ABC, ABCD\}$), all sharing the same first
constituent.
\textbf{(b)}~The mirror image---a suffix-closed dictionary
$\{CD, BCD, ABCD\}$---never arises under LZW.
\textbf{(c)}~Consequence for the language model: at a single decoding position
the output hyper-encoder must score every member of such a chain against the
same hidden state, which pushes their unembedding vectors toward one another
and imposes a prefix-aligned geometry on $f_\phi^{\text{out}}$.}
\label{fig:prefix-suffix}
\end{figure}

The key insight is that the LZW dictionary is \emph{prefix-closed}: a hypertoken
$z$ with constituents $x_{1:K}$ enters $\mathcal{V}_h$ only after $x_{1:K-1}$ has
already been registered, so $\mathcal{V}_h$ is precisely the node set of a trie
over base-token strings.
This is the maximal violation of the prefix-free condition---every internal node
is a prefix of some other entry---and hence $\mathcal{V} \cup \mathcal{V}_h$ is
not a uniquely decipherable code: the map from segmentations to base-token
strings is many-to-one (\S\ref{sec:multi-view}). Standard LZW appears
deterministic only because its encoder imposes a greedy longest-match parsing
rule; once the encoder is a sampling language model, that tie-break disappears
and hypertokens sharing a prefix genuinely compete at the same decoding position, as illustrated in \Cref{fig:prefix-suffix}(c).
For the output hyper-encoder this means $v_{\text{out}}(x_{1:K})$ and
$v_{\text{out}}(x_{1:K-1})$ are scored against one hidden state, which favours a
strongly \emph{prefix-aligned} geometry on $f_\phi^{\text{out}}$.
The input hyper-encoder $f_\phi^{\text{in}}$, in contrast, faces no such
structural constraint. Its role is to produce an embedding $v(z)$
that best serves to predict the next token---which, as we show empirically, leads
to a \emph{balanced} geometry that weighs every constituent of $x_{1:K}$
comparably. These are opposing objectives---one geometry is prefix-dominated,
the other treats all positions alike---and we hypothesise that one shared
encoder cannot serve both well.

\paragraph{Decoupling Input and Output Hypertoken Embeddings.}
In conventional transformer language models, tying the input embedding matrix
$\mathbf{W}_{\text{in}}$ and the output unembedding matrix $\mathbf{W}_{\text{out}}$
is standard practice~\citep{press-wolf-2017-using, inan-etal-2017-tying}, particularly
for smaller models. This reduces parameters without degrading performance, as both
matrices operate over the same fixed vocabulary $\mathcal{V}$ and each token serves
a single, static role.
For \texttt{zip2zip} one might similarly wish to tie $f_\phi^{\text{in}}$ and
$f_\phi^{\text{out}}$. Due to the conflict above, however, a shared encoder would
have to serve both geometries at once and ends up serving neither. We therefore
untie the two hyper-encoders, at the cost of one additional lightweight encoder.

\paragraph{Experiments.}
We verify the above hypothesis by probing the learned geometry of the untied
hyper-encoders ($f_\phi^{\text{in}}$ and $f_\phi^{\text{out}}$) after training.
\emph{\textbf{Substitution probe.}} Take a hypertoken with constituents $ABC$ and
perturb exactly one end: replacing the first constituent gives $KBC$, replacing
the last gives $ABJ$, where $J$ and $K$ are drawn uniformly from $\mathcal{V}$.
We compute the input and output embeddings of the original and perturbed hypertokens, and measure the cosine similarity between them. 
We then obtain the ratio $r_K = \cos_{ABC, ABJ} / \cos_{ABC, KBC}$, which
quantifies positional asymmetry: $r_K > 1$ indicates \textbf{prefix-aligned}
representations, $r_K < 1$ \textbf{suffix-aligned}, and $r_K \!\approx\! 1$ a
\textbf{balanced} representation that weighs both ends equally.

\begin{figure}[h]
\centering
\includegraphics[width=\textwidth]{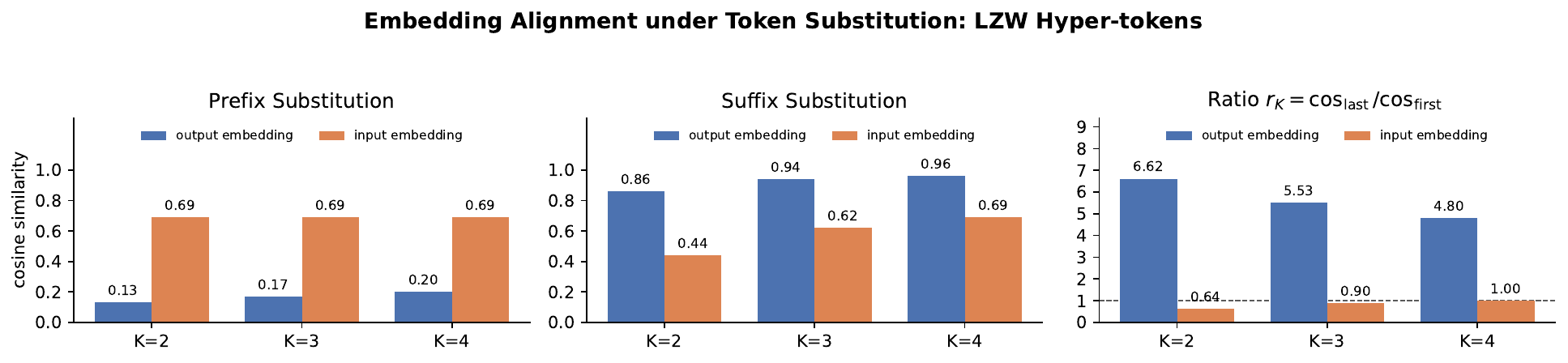}
\caption{\textbf{Substitution probe on LZW hypertokens} (zip2zip++, Phi-3.5-mini). Bars report the
mean cosine \emph{similarity} between a hypertoken embedding and the same
embedding after replacing its first (\emph{Prefix Substitution}) or last
(\emph{Suffix Substitution}) constituent. The ratio $r_K = \cos_{\text{last}}/\cos_{\text{first}}$ (right) quantifies
positional asymmetry. The output hyper-encoder is strongly prefix-dominated at
every span length, whereas the input hyper-encoder is
balanced. The
corresponding BPE control, which shows no comparable asymmetry, is
deferred to \Cref{fig:substitution-barplot-bpe}.}
\label{fig:substitution-barplot}
\end{figure}

\paragraph{Observations.} Three findings emerge from the substitution probe.
\textbf{(1)~The output encoder is prefix-aligned:} replacing the first
constituent moves the embedding far more than replacing the last.
\textbf{(2)~The input encoder is balanced:} its substitution effects are similar at both ends.
\textbf{(3)~Neither pattern is present in the base BPE embeddings:} running the
same experiments on the base BPE embeddings shows no comparable asymmetry, indicating that these patterns are learned from LZW compressed sequences rather than being general properties of BPE embeddings.
As a second check we vary the \emph{extent} of a
span rather than perturbing it in place: given a span $ABCD$, we embed every
contiguous sub-span and compare the nested prefix chain
($AB \subset ABC \subset ABCD$) against the sub-spans that drop $A$ (we refer to this as \emph{Sequential Probe}). The
resulting similarity structure (\Cref{fig:sequence-heatmap}) further supports the
conclusion above---the output embedding clusters by shared prefix, whereas the
input embedding tracks total constituent overlap. Details are deferred to
\Cref{app:embedding_probes}.

\subsection{Residual Hyper-Encoder with Identity Initialization}
\label{sec:residual}

\begin{figure}[h]
    \centering
    \begin{subfigure}[b]{0.35\textwidth}
        \centering
        \includegraphics[width=\linewidth]{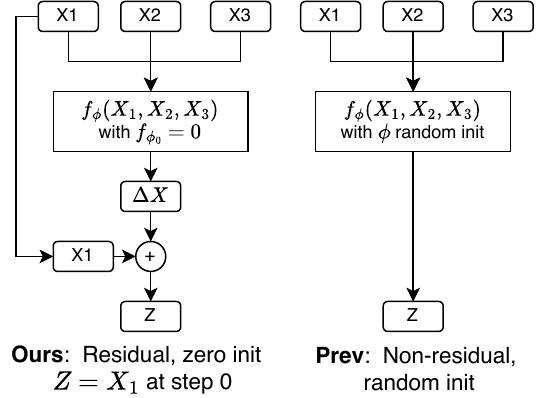}
        \caption{}
        \label{fig:residual-diagram}
    \end{subfigure}
    \hfill
    \begin{subfigure}[b]{0.64\textwidth}
        \centering
        \includegraphics[width=\linewidth]{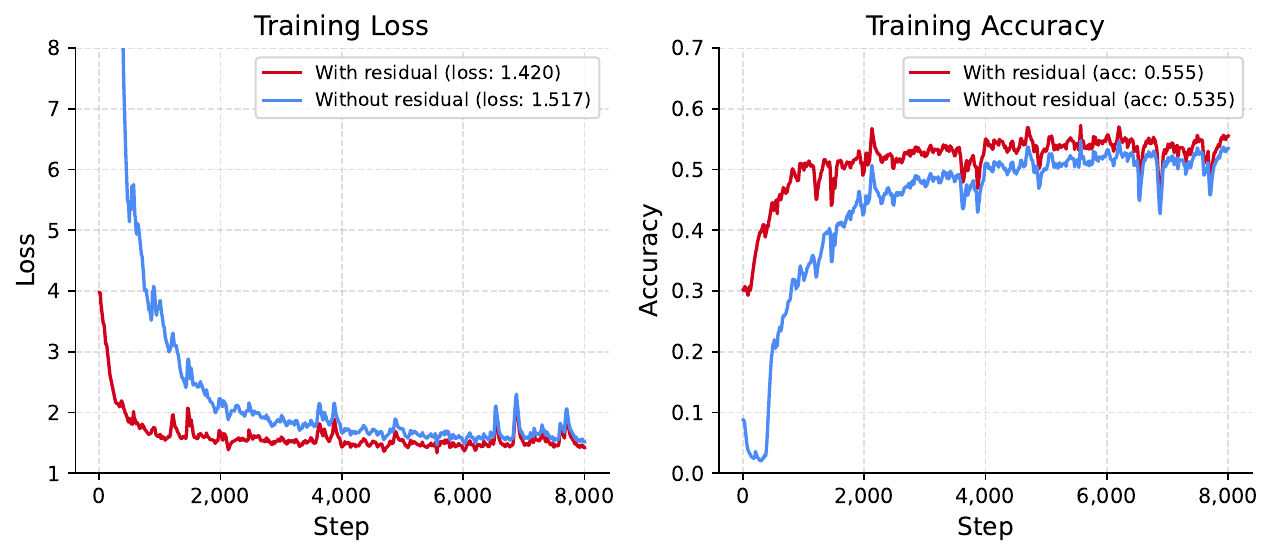}
        \caption{}
        \label{fig:residual-curves}
    \end{subfigure}
    \caption{\textbf{Identity initialization of the hyper-encoder.}
    \textbf{(a)}~With the residual, the encoder only predicts a correction
    $\Delta X$ on top of $e_{x_1}$ and is zero-initialized ($f_{\phi_0}=0$), so a
    hypertoken embedding starts out exactly at $e_{x_1}$.
    \textbf{(b)}~Training dynamics: identity initialization removes the early
    loss shock and reaches both a better final loss and a better accuracy.}
    \label{fig:residual-ablation}
\end{figure}

\paragraph{Problem: Hypertoken Embedding Scale Mismatch.}
In the original \texttt{zip2zip}, a hypertoken embedding is read straight off a
randomly initialized encoder, $v(z) = f_\phi(e_{x_1}, \dots, e_{x_K})$, where
$e_{x_k} := \mathbf{W}_{\text{in}}[x_k]$. The scale mismatch this creates is
severe. The encoder terminates in a LayerNorm, which rescales every coordinate to unit variance. Its output vectors therefore have norm ${\approx}\sqrt{d} \approx 55$, roughly $54\times$ the norm of a base token embedding. Every hypertoken therefore enters the decoder as a large out-of-distribution vector, and
the resulting optimization shock is severe: step-10 loss of $70$--$100$, off the scale of \Cref{fig:residual-curves}.

\paragraph{Residual Hyper-Encoder with Identity Initialization.}
We instead give the encoder a residual connection onto the embedding of the
\emph{first} constituent base token (\Cref{fig:residual-diagram}):
\begin{equation}
    v(z) = e_{x_1} + f_\phi(e_{x_1}, \dots, e_{x_K}),
    \qquad f_{\phi_0} = 0,
    \label{eq:residual-hyper-encoder}
\end{equation}
where $f_\phi$ denotes the encoder function.
 We enforce $f_{\phi_0}=0$ by setting the
scale $\gamma$ and bias $\beta$ of the encoder's final LayerNorm to zero. This
guarantees $v(z) = e_{x_1}$ exactly at the first step, so every hypertoken enters the
decoder as a genuine pretrained token embedding and the encoder only has to learn
how to \emph{move away} from it. This is the trick LoRA uses---zero initializing
$B$ in $\Delta W = BA$ so that adaptation departs from the pretrained function
rather than from noise---applied to the embedding table instead of the weights.
As shown in \Cref{fig:residual-ablation}, this simple change has a dramatic effect on training dynamics.
It removes the shock above (step-10 loss $3.97$, gradient norm ${\approx}50$) and
yields a better final loss and accuracy.

\subsection{Size-aware RoPE}
\label{sec:rope}

\begin{figure}[h]
    \centering
    \begin{subfigure}[b]{0.372\textwidth}
        \includegraphics[width=\linewidth]{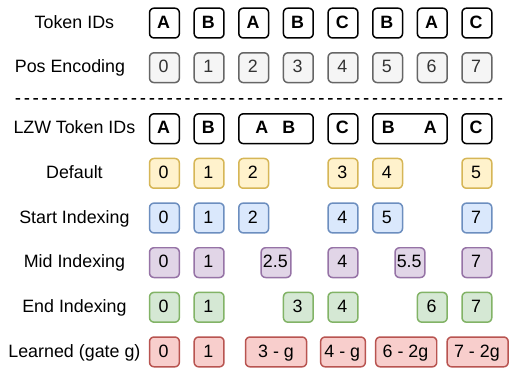}
        \caption{Size-aware RoPE}
        \label{fig:methods-rope}
    \end{subfigure}\hspace{0.004\textwidth}%
    \begin{subfigure}[b]{0.321\textwidth}
        \includegraphics[width=\linewidth]{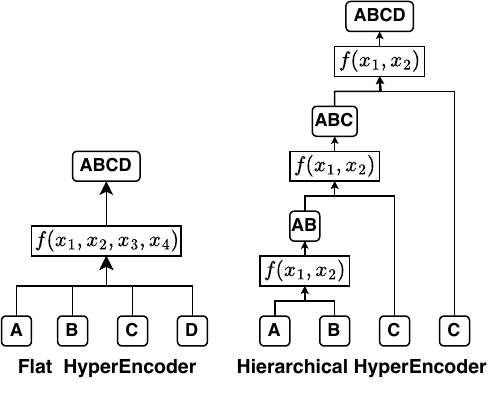}
        \caption{Hierarchical hyper-encoder}
        \label{fig:methods-hier}
    \end{subfigure}\hspace{0.004\textwidth}%
    \begin{subfigure}[b]{0.281\textwidth}
        \includegraphics[width=\linewidth]{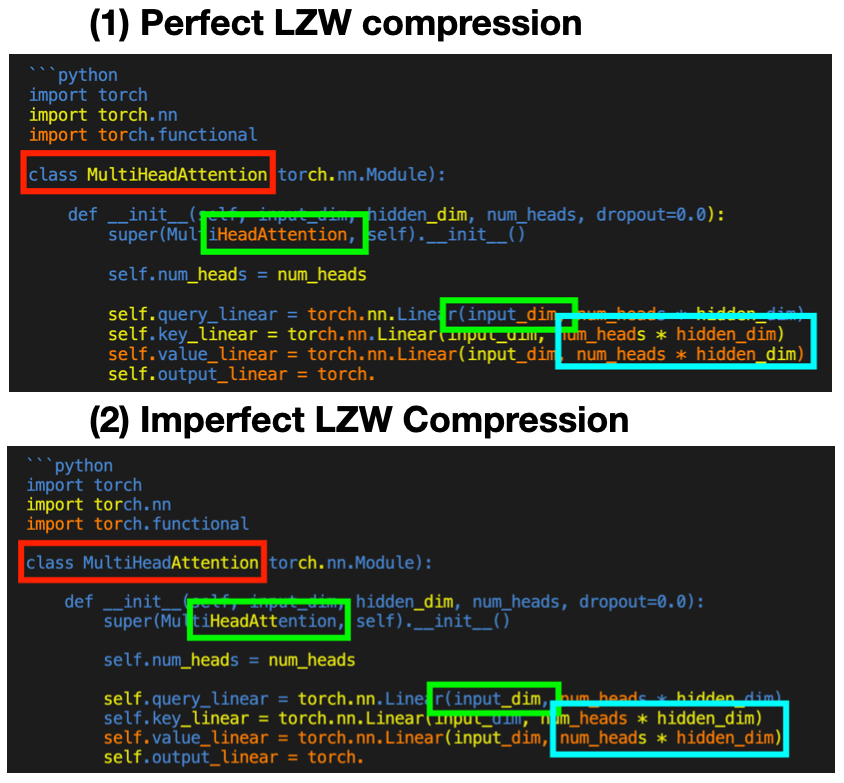}
        \caption{Imperfect compression}
        \label{fig:methods-imperfect}
    \end{subfigure}
    \caption{
    \textbf{(a)}~Position assignment for a compressed sequence
    (\Cref{sec:rope}). The base tokens $ABABCBAC$ occupy positions $0$--$7$; LZW
    merges them into $A$, $B$, $AB$, $C$, $BA$, $C$. \emph{Default} numbers the
    compressed tokens consecutively;
    \emph{start}, \emph{mid} and \emph{end} indexing take the first, mean and last
    base position of each span and leave the other base tokens in
    place. The \emph{learned} rule interpolates between end indexing and the
    default index with a gate $g$ on the low-frequency rotary pairs ($g=0$ is end
    indexing, $g=1$ the default). 
    \textbf{(b)}~Two hyper-encoder variants (\Cref{sec:merge-size-transfer}).
    \emph{Flat} encodes all constituents tokens in a single pass and is tied to the
    training-time merge size; \emph{hierarchical} folds them pairwise with a shared
    composition function.
    \textbf{(c)}~Perfect versus imperfect LZW compression of the same code snippet
    (\Cref{sec:lzw}). Coloured boxes mark different hypertokens.}
    \label{fig:methods}
\end{figure}

Merging tokens shortens the sequence, which raises the question of what RoPE
position a hypertoken should receive. In the original implementation,
\citet{geng2025zip2zip} simply number the compressed tokens $0, 1, 2, \ldots$,
treating them the same as base tokens. One consequence is that the RoPE
coordinates of every subsequent token are shifted by $K-1$ for each merged span of
length $K$, as shown in \Cref{fig:methods-rope}.
Here we instead assign a hypertoken a position that accounts for its size, which
we call \emph{size-aware RoPE}. Since a hypertoken $z$ with constituents $x_{1:K}$
occupies the span of base tokens $x_1, \dots, x_K$, a natural choice is to reuse
one of those base positions. We consider three variants---\emph{start},
\emph{mid} and \emph{end} indexing---which take the first, mean and last base
position of the span.
As \Cref{fig:methods-rope} shows, the positions of the remaining base tokens are
then unaffected by the merge: their absolute coordinates, and the relative
distances between them, are the same as in the uncompressed sequence. This should
reduce the distortion that compression introduces into the RoPE geometry.
In addition, we also consider a \emph{learned} rule that interpolates between the end-indexed position and the default compressed index, $p = p_{\text{end}} + g\,(p_{\text{comp}} - p_{\text{end}})$, with one gate $g$ per layer and rotary frequency pair on the low-frequency pairs, initialised at zero (end indexing) and learned during finetuning.

\Cref{tab:ablation} supports this: adding end indexing to the digit-protected,
untied model raises GSM8K by 2.7 points at unchanged perplexity. The learned rule
matches its perplexity but not its reasoning, so we keep the parameter-free end
indexing (\Cref{app:negative_results}).

\subsection{Hierarchical Hyper Encoder}
\label{sec:merge-size-transfer}

The hyper-encoder described so far is \emph{flat}: it takes all $K$ constituents
at once, $f_\phi(e_{x_1},\dots,e_{x_K})$, padding shorter spans up to the maximum
merge size $M$ (\Cref{fig:methods-hier}, left). This gives direct access to the
whole constituent sequence, but ties the encoder's learned positional structure to
the $M$ used during training, so a larger merge size at evaluation time is out of
distribution.

We therefore also consider a \emph{hierarchical} hyper-encoder, which builds the
same embedding by repeatedly applying $f_\phi$ to \emph{two} arguments
(\Cref{fig:methods-hier}, right): for constituents $[A,B,C,D]$ it first composes
$A$ and $B$ into $[A,B]$, then $[A,B]$ and $C$ into $[A,B,C]$, and finally
$[A,B,C]$ and $D$ into $[A,B,C,D]$, each step reusing the same weights. Because the pairwise composer is
shared across all depths, increasing the evaluation-time merge size adds recursion
steps but no new parameters and no length-specific embeddings---the property we
test in \Cref{app:merge_size_transfer}.

\section{A Fault-Tolerant LZW Codec}

The original zip2zip~\citep{geng2025zip2zip} adapted the classical LZW codec into token space, but the algorithm is the same. 
Yet, Language Models (LMs), which generate tokens autoregressively, have some distinct properties that make the original LZW algorithm (designed for text compression) brittle.

\begin{figure}[H]
    \centering
    \includegraphics[width=\linewidth]{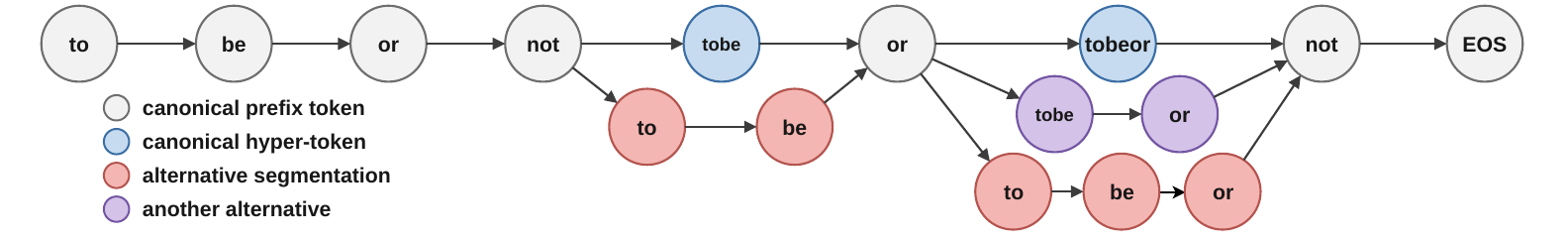}
    \caption{%
        \textbf{Segmentation tree for multi-view scoring.}
        Segmentation tree for the text
        \texttt{to be or not to be or to be or not}. The canonical chain (grey and
        blue) carries the greedy LZW segmentation;  all
        views of a position decode to the same base-token string.%
    }
    \label{fig:segmentation-tree}
\end{figure}

\paragraph{Multi-View Segmentation Problem.}
\label{sec:multi-view}
A fixed BPE vocabulary already admits many tokenizations of the same text and merely picks one by merge order, as shown in Figure~\ref{fig:segmentation-tree}, so scoring only that one understates a model's likelihood~\citep{cao-rimell-2021-marginal}---though for BPE the correction is small, under $0.5\%$ of log-likelihood in most cases~\citep{chirkova-etal-2023-marginalize}. In \texttt{zip2zip} the LZW-induced vocabulary makes this \emph{tokenization ambiguity} central: a single string can be segmented into hypertokens in many valid ways, all of which decompress to the same output (see Figure~\ref{fig:methods-imperfect} for an example). We refer to the set of equivalent segmentations as the \emph{multi-view} of a given string.
Deterministic LZW hides this multiplicity behind a greedy longest-match rule, which selects one \emph{canonical} segmentation; a language model trained in compressed space has no such guarantee.
During training, the model is exposed to the canonical LZW segmentation, but at inference time it may emit any valid alternative. 
While the generated text remains correct upon decompression, this distributional spread poses a challenge for evaluation: standard metrics that condition on a single canonical segmentation will underestimate the model's true performance.

\paragraph{Recovering from missed merges.}
\label{sec:lzw}
Standard LZW decompression assumes a perfectly valid compressed
sequence---a property that holds for deterministic compressors but breaks
when the compressor is an autoregressive neural model, as demonstrated in
Figure~\ref{fig:methods-imperfect}.
Any under-merging violates the one-to-one correspondence between received
codes and novel dictionary entries, causing codebook desynchronization and
cascading errors.
The main change FT-LZW makes to the codec is to replace the codec's single unconditional
insertion with a \emph{guarded merge loop}
(Algorithm~\ref{alg:lzw-contrast}).
Upon decoding code~$c_t$ to sequence~$\omega$, the decoder walks along $\omega$,
extending the buffer~$W$ one symbol at a time: if $W \!\cdot\! y$ is already in the
dictionary---the signature of a merge the model failed to make---the symbol is
absorbed into~$W$ and the next symbol of~$\omega$ is tried; an entry is inserted
only when a genuinely novel combination is reached.
On perfectly compressed input the guard never fires and FT-LZW recovers exactly the
dictionary of standard LZW; on imperfect input it degrades gracefully to emitting
base tokens rather than producing garbage.

\paragraph{Protecting fragile tokens.}
Standard LZW codecs blindly merge every pair of tokens they encounter, regardless of semantics. 
For LMs, however, certain tokens should not be merged. 
This applies to at least two cases: (1) special tokens used to delimit system, user, and assistant messages or signal EOS, and 
(2) tokens where merging is known to hurt performance, such as digits~\citep{singh2024tokenizationcountsimpacttokenization}. 
To handle this, we add a short-circuit to the codec's compression loop: whenever a token from the protected set is encountered, we emit it directly as a base token and skip the merge step.

The full codec---both the batch decoder and the online variant used during
generation---is given in \Cref{app:ftlzw}, and we release it as an open-source Rust
library.\footnote{\url{https://github.com/epfl-dlab/zip2zip-compression}}

\subsection{Multi-View Perplexity}
\label{sec:fault-tolerant}

To account for this ambiguity, we define a \emph{multi-view} likelihood that aggregates over equivalent complete segmentations at the probability level. Concretely, for each target hypertoken $z_i$, let $\mathcal{S}(z_i)$ denote the set of all valid segmentations of $\mathrm{expand}(z_i)$---i.e., token sequences that decode to the same base-token string. The quantity sums the probability mass assigned to all equivalent segmentations:
\begin{equation}
    p_{\text{multi-view}}(z_i\mid z_{<i}) = \sum_{s \in \mathcal{S}(z_i)} p(s \mid z_{<i}) = \sum_{s \in \mathcal{S}(z_i)} \prod_{j=1}^{|s|} p(s_j \mid z_{<i}, s_{<j}).
    \label{eq:multi-view}
\end{equation}
Intuitively, this captures the total probability the model assigns to \emph{any} valid way of producing the same output text, rather than penalizing it for choosing a non-canonical but equally correct segmentation. For base tokens, $\mathcal{S}(z_i)$ reduces to the singleton $\{z_i\}$, and the multi-view probability coincides with the standard one.
The multi-view perplexity replaces the token probability with this aggregate over the $N$ canonical tokens (byte-level scores divide the same sum by the byte count of $x$):
  \begin{equation}
      \mathrm{PPL}_{\text{multi-view}}(x)
      = \exp\!\left(
          -\frac{1}{N}\sum_{i=1}^{N}
          \log\, p_{\text{multi-view}}(z_i \mid z_{<i})
        \right).
  \end{equation}

For a hypertoken expanding to $n$ base tokens, the number of valid
segmentations $|\mathcal{S}(z_i)|$ grows as $\mathcal{O}(2^n)$, but the merge cap $M$ bounds it by $2^{M-1}$,
which stays small in practice. Using sequence packing and a tree-structured attention mask, we evaluate Eq.~\eqref{eq:multi-view} exactly, given the canonical prefix, in a single additional forward pass at only $1.6\times$ the cost of standard perplexity (details in \Cref{app:tree_attention}).

\section{Experiments}
\label{sec:experiments}

\paragraph{Baselines.}
We compare three recipes that share one finetuning configuration. \texttt{Uncompressed} is the uncompressed control: the same
base model, data, LoRA configuration and update budget, with compression disabled,
so it isolates what our data mixture does on its own. \texttt{zip2zip} is the
original recipe of~\citet{geng2025zip2zip}. \texttt{zip2zip++} adds the components of
this paper on top of it: protected digits (\Cref{sec:lzw}), untied hyper-encoders
(\Cref{sec:untied}), size-aware RoPE (\Cref{sec:rope}), the identity-initialised
residual hyper-encoder (\Cref{sec:residual}).

\paragraph{Models and data.}
We evaluate on four instruction-tuned decoders: Phi-3.5-mini-instruct (3.8B) and
Phi-3-medium-4k-instruct (14B) in the main text, and Llama-3.2-1B/3B-Instruct in
\Cref{tab:nlp_perplexity_full}. All four are LoRA finetunes for 8{,}000 steps on
the 1B-finetuning data released in \citet{geng2025zip2zip}.
The maximum merge size is $M{=}4$ throughout. Full hyperparameters are in
\Cref{app:experiment_setup}.

\paragraph{Metrics.}
We report three families. \emph{Accuracy} on six two-shot multiple-choice
benchmarks and, after finetuning, on GSM8K, MATH-500, HumanEval, IFEval and RULER;
per-benchmark details are in \Cref{app:benchmarks}. \emph{Byte-level perplexity}
on four corpora, in two variants: \emph{strict}, which scores only the canonical
LZW segmentation of each target, and \emph{multi-view} ($\star$), which sums the
probability of every valid segmentation (\Cref{sec:fault-tolerant}); byte
normalisation makes the two vocabularies comparable.

\begin{table*}[htbp]
    \centering
    \small
    \setlength{\tabcolsep}{3.5pt}
    \caption{\small \textbf{Main results.} Two-shot accuracy on six benchmarks and
    byte-level perplexity ($\downarrow$) on four corpora with a 1024-token context;
    details in \Cref{app:benchmarks}. \textbf{Uncompressed} is the same base model
    finetuned on uncompressed sequences under the same conditions as in
    \Cref{app:experiment_setup}. Bold marks the better of the two compressed models
    in each column. Results for Llama-3.2-1B/3B-Instruct are in \Cref{tab:nlp_perplexity_full}.}
    \label{tab:nlp_perplexity}
    \begin{tabular}{ll|cccccc|cccc}
        \toprule
        \textbf{Model} & \textbf{Method}
        & {ARC-c} & {ARC-e} & {HS} & {OBQA} & {PIQA} & {WG}
         & {Wiki} & {Pile} & {mC4} & {dC4} \\
        \midrule
        \multirow{3}{*}{Phi-3.5-4B}
        & Uncompressed  & 0.59 & 0.84 & 0.74 & 0.48 & 0.82 & 0.75 & 1.58 & 1.67 & 1.77 & 1.82 \\
        & zip2zip   & 0.54 & 0.81 & 0.71 & 0.46 & \textbf{0.80} & \textbf{0.75} & 1.70 & 1.82 & 1.86 & 1.92 \\
        & zip2zip++ & \textbf{0.57} & \textbf{0.83} & \textbf{0.72} & \textbf{0.47} & \textbf{0.80} & 0.74 & \textbf{1.66} & \textbf{1.78} & \textbf{1.83} & \textbf{1.89} \\
        \hline
        \multirow{3}{*}{Phi-3-14B}
        & Uncompressed  & 0.65 & 0.88 & 0.81 & 0.51 & 0.83 & 0.79 & 1.43 & 1.61 & 1.71 & 1.76 \\
        & zip2zip   & 0.61 & 0.85 & 0.78 & \textbf{0.51} & 0.81 & 0.77 & 1.58 & 1.76 & 1.80 & 1.85 \\
        & zip2zip++ & \textbf{0.62} & \textbf{0.86} & \textbf{0.79} & 0.50 & \textbf{0.82} & \textbf{0.78} & \textbf{1.53} & \textbf{1.72} & \textbf{1.77} & \textbf{1.82} \\
        \bottomrule
    \end{tabular}
\end{table*}

\begin{table*}[htbp]
    \centering
    \small
    \setlength{\tabcolsep}{3.5pt}
    \caption{\small Perplexity evaluation on four corpora using a 1024-token context
    window. $\star$ denotes the exact multi-view variants of
    \Cref{sec:fault-tolerant}, which sum the probability of every valid segmentation
    of the target. The uncompressed reference has a unique tokenization, so its
    multi-view scores coincide with its strict ones. PPL is per whitespace-delimited
    word, BPB per byte.}
    \label{tab:perplexity}
    \resizebox{\textwidth}{!}{
    \begin{tabular}{ll|cccc|cccc|cccc|cccc}
        \toprule
        & & \multicolumn{4}{c|}{Wiki} & \multicolumn{4}{c|}{Pile} & \multicolumn{4}{c|}{mC4} & \multicolumn{4}{c}{dC4} \\
        \cmidrule(lr){3-6} \cmidrule(lr){7-10} \cmidrule(lr){11-14} \cmidrule(lr){15-18}
        \textbf{Model} & \textbf{Training}
        & PPL & PPL$^\star$ & BPB & BPB$^\star$
        & PPL & PPL$^\star$ & BPB & BPB$^\star$
        & PPL & PPL$^\star$ & BPB & BPB$^\star$
        & PPL & PPL$^\star$ & BPB & BPB$^\star$ \\
        \midrule
        \multirow{3}{*}{Phi-3.5-4B}
        & Uncompressed & 11.5 & 11.5 & 0.66 & 0.66 & 31.5 & 31.5 & 0.74 & 0.74 & 30.4 & 30.4 & 0.82 & 0.82 & 35.2 & 35.2 & 0.87 & 0.87 \\
        & zip2zip      & 16.8 & 14.8 & 0.76 & 0.73 & 57.9 & 49.5 & 0.87 & 0.83 & 41.0 & 37.3 & 0.89 & 0.87 & 47.5 & 43.0 & 0.94 & 0.91 \\
        & zip2zip++    & \textbf{14.9} & \textbf{13.6} & \textbf{0.73} & \textbf{0.70} & \textbf{49.3} & \textbf{43.4} & \textbf{0.83} & \textbf{0.81} & \textbf{37.8} & \textbf{35.0} & \textbf{0.87} & \textbf{0.85} & \textbf{44.0} & \textbf{40.7} & \textbf{0.92} & \textbf{0.90} \\
        \hline
        \multirow{3}{*}{Phi-3-14B}
        & Uncompressed &  6.8 &  6.8 & 0.52 & 0.52 & 25.0 & 25.0 & 0.69 & 0.69 & 24.9 & 24.9 & 0.77 & 0.77 & 28.6 & 28.6 & 0.81 & 0.81 \\
        & zip2zip      & 11.6 &  9.9 & 0.66 & 0.62 & 45.4 & 38.7 & 0.82 & 0.78 & 33.6 & 30.5 & 0.84 & 0.82 & 38.5 & 34.9 & 0.89 & 0.86 \\
        & zip2zip++    & \textbf{9.7} & \textbf{8.8} & \textbf{0.61} & \textbf{0.59} & \textbf{39.0} & \textbf{34.7} & \textbf{0.78} & \textbf{0.76} & \textbf{30.7} & \textbf{28.6} & \textbf{0.82} & \textbf{0.81} & \textbf{35.5} & \textbf{33.0} & \textbf{0.87} & \textbf{0.85} \\
        \bottomrule
    \end{tabular}
    }
\end{table*}

\paragraph{Evaluation I: Perplexity and Few-Shot Accuracy.}
\Cref{tab:nlp_perplexity} reports six two-shot benchmarks together with strict
byte-level perplexity, which scores only the canonical LZW segmentation of each
target; \Cref{tab:perplexity} adds the multi-view score.
\textbf{(1) \texttt{zip2zip++} improves on \texttt{zip2zip} across the board.}
Perplexity is lower on all four corpora at both scales, under strict and multi-view
scoring alike, and accuracy matches or beats \texttt{zip2zip} in most tasks.
\textbf{(2) The gap to uncompressed is roughly halved, not closed.} Of the strict
gap \texttt{zip2zip} leaves, the recipe closes about a third and multi-view scoring
a further sixth to a third (a scoring artefact, not a modelling failure); the rest is
unattributed, as the score conditions on the canonical prefix (\Cref{app:tree_attention}).

\begin{table}[h]
    \centering
    \small
    \caption{\textbf{Post-SFT evaluation} on generation tasks and long context. All
    three recipes share the same data stream and update budget and differ only in the
    training representation and its window filter (\Cref{app:variance_comparability});
    per-benchmark details are in \Cref{app:benchmarks} and
    strict-format variants in \Cref{app:answer_extraction}. Bold marks the better of
    the two compressed models in each column; \textbf{Uncompressed} is the reference.}
    \label{tab:post-sft}
    \vspace{0.5em}
    \setlength{\tabcolsep}{3.5pt}

    \begin{tabular}{ll c c c c c c c c}
        \toprule
        &
        & \textbf{GSM8K}
        & \textbf{MATH-500}
        & \textbf{HumanEval}
        & \textbf{IFEval}
        & \multicolumn{4}{c}{\textbf{RULER}} \\

        \cmidrule(lr){3-3}
        \cmidrule(lr){4-4}
        \cmidrule(lr){5-5}
        \cmidrule(lr){6-6}
        \cmidrule(lr){7-10}

        \textbf{Scale}
        & \textbf{Model}
        & acc
        & acc
        & pass@1
        & prompt acc
        & 4K
        & 8K
        & 16K
        & 32K \\

        \midrule

        \multirow{3}{*}{Phi-3.5-4B}
        & Uncompressed  & 0.75 & 0.36 & 0.58 & 0.38 & 86.4 & 81.5 & 78.1 & 73.7 \\
        & zip2zip   & 0.36 & 0.15 & 0.26 & \textbf{0.32} & 46.6 & 28.1 & 21.0 & 15.2 \\
        & zip2zip++ & \textbf{0.68} & \textbf{0.24} & \textbf{0.32} & \textbf{0.32} & \textbf{70.7} & \textbf{60.6} & \textbf{53.2} & \textbf{52.8} \\
        \midrule
        \multirow{3}{*}{Phi-3-14B}
        & Uncompressed  & 0.70 & 0.39 & 0.57 & 0.49 & 81.6 & 47.9 & 40.6 & 32.2 \\
        & zip2zip   & 0.51 & 0.22 & \textbf{0.36} & \textbf{0.43} & 52.5 & 23.2 & 14.0 & 11.6 \\
        & zip2zip++ & \textbf{0.69} & \textbf{0.30} & 0.31 & 0.40 & \textbf{69.2} & \textbf{30.3} & \textbf{24.0} & \textbf{17.1} \\
        \bottomrule
    \end{tabular}
\end{table}

\paragraph{Evaluation II: Downstream Tasks.}
\Cref{tab:post-sft} evaluates on tasks that require generation and reasoning.
\textbf{(1) Mathematical reasoning improves the most.} GSM8K rises from $0.36$ to
$0.68$ at 3.8B and from $0.51$ to $0.69$ at 14B, the latter within a point of
\texttt{Uncompressed} ($0.70$). MATH-500 improves from $0.15$ to $0.24$ and from $0.22$
to $0.30$, but stays well below \texttt{Uncompressed} ($0.36$ and $0.39$).
\textbf{(2) Code generation and instruction following are not improved.} At 3.8B
HumanEval moves from $0.26$ to $0.32$ and IFEval is unchanged at $0.32$; at 14B
\texttt{zip2zip++} is \emph{worse} than \texttt{zip2zip} on both.

\paragraph{Evaluation III: Long-Context Understanding.}
The RULER columns of \Cref{tab:post-sft} probe retrieval at 4K--32K context.
\textbf{(1) \texttt{zip2zip++} recovers a large share of what compression costs.}
At 3.8B it raises 4K accuracy from $46.6$ to $70.7$ and 32K from $15.2$ to $52.8$,
against $86.4$ and $73.7$ for \texttt{Uncompressed}; the same ordering holds at 14B.
\textbf{(2) The absolute numbers are limited by the training length, not by
compression alone.} None of the three recipes received long-context training: they
were finetuned with 2{,}048-token sequences and evaluated with inference-time
LongRoPE adaptation (\Cref{app:benchmarks}).

\begin{table*}[t]
    \centering
    \small
    \setlength{\tabcolsep}{2.5pt}
    \caption{\small \textbf{Ablation study on Phi-3.5-mini.} Rows are cumulative, in the order in which the recipe was developed; the last row of the ladder is the zip2zip++ model of \Cref{tab:nlp_perplexity}. Two-shot accuracy and byte-level perplexity $\downarrow$, as in \Cref{tab:nlp_perplexity}. The continued-pretraining row is a single run after 200B additional tokens of compressed pretraining, not compute-matched to the others. Per-component gains and paired significance tests are in \Cref{app:ablation_stats}. Bold marks the best value along the ladder; the rows below the rule are references.}
    \label{tab:ablation}
    \begin{tabular}{p{3.3cm}|ccccccc|cccc}
        \toprule
        \textbf{Configuration} 
        & {ARC-c} & {ARC-e} & {HS} & {OBQA} & {PIQA} & {WG} & {GSM8K} 
        & {Wiki} & {Pile} & {mC4} & {dC4} \\
        \midrule
        zip2zip & 0.54 & 0.81 & 0.71 & 0.46 & \textbf{0.80} & \textbf{0.75} & 0.36 & 1.70 & 1.82 & 1.86 & 1.92 \\
        + Digit protection & 0.54 & 0.81 & 0.70 & \textbf{0.47} & 0.79 & 0.71 & 0.61 & 1.71 & 1.83 & 1.87 & 1.93 \\
        + Untied hyper-encoders & 0.56 & 0.82 & \textbf{0.72} & \textbf{0.47} & 0.79 & 0.74 & 0.62 & 1.69 & 1.81 & 1.85 & 1.91 \\
        + Size-aware RoPE & 0.55 & 0.82 & 0.71 & 0.46 & \textbf{0.80} & 0.74 & 0.65 & 1.69 & 1.81 & 1.85 & 1.91 \\
        + Identity-init residual  & \textbf{0.57} & \textbf{0.83} & \textbf{0.72} & \textbf{0.47} & \textbf{0.80} & 0.74 & \textbf{0.68} & \textbf{1.66} & \textbf{1.78} & \textbf{1.83} & \textbf{1.89} \\
        \midrule
        + Continued Pretraining  & 0.55 & 0.82 & 0.72 & 0.46 & 0.80 & 0.75 & 0.64 & 1.63 & 1.74 & 1.82 & 1.88 \\
        \bottomrule
    \end{tabular}
\end{table*}

\paragraph{Evaluation IV: Ablation Study.}
\Cref{tab:ablation} adds one component at a time, in the order the recipe was
developed.
\textbf{(1) Protecting digits dominates.} It alone moves GSM8K from $0.36$ to
$0.61$; the three architectural changes add the remaining $7$ points, with
size-aware RoPE and the identity-initialised residual contributing $+2.7$ and $+2.8$.
Untying the hyper-encoders acts mainly on the other metrics ($+1.6$ points of mean
multiple-choice accuracy, $-0.025$ WikiText byte-perplexity).
\textbf{(2) The end-to-end effect is significant; the individual rungs are less so.}
Paired significance tests against both \texttt{zip2zip} and \texttt{Uncompressed},
together with the seed spread, are reported in \Cref{app:ablation_stats}.
\textbf{(3) More compressed pretraining improves perplexity but not reasoning.}
A further 200B tokens of compressed pretraining gives the best perplexity in the
table ($1.63$ on WikiText) while GSM8K drops from $0.68$ to $0.64$---one more case
where perplexity and reasoning move apart.

\begin{figure}[H]
    \centering
    \includegraphics[width=\textwidth]{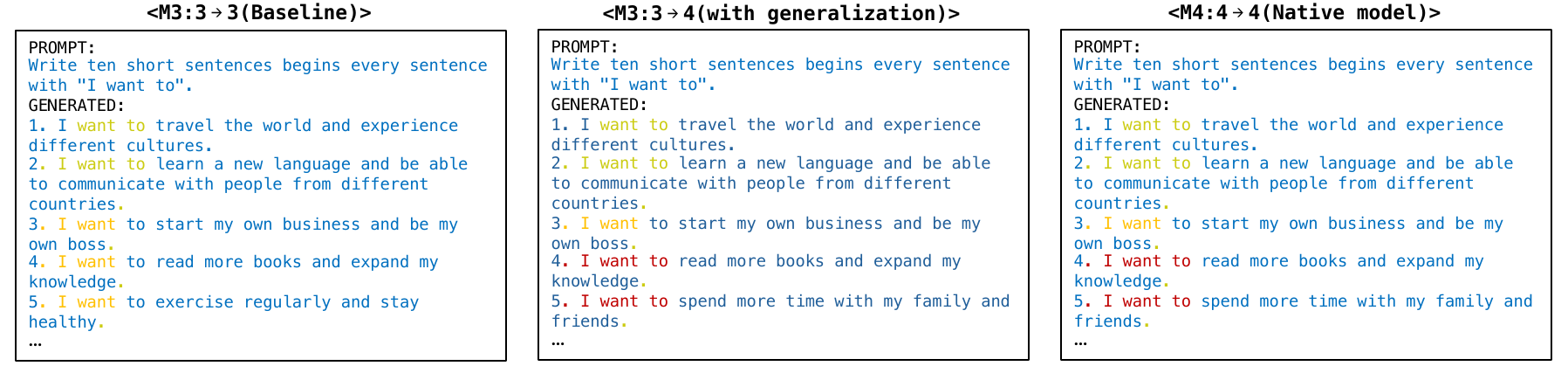}
    \caption{
        \textbf{Hypertoken length generalization examples.}
        Left: $M_3$ evaluated with its training-time maximum merge size
        ($3{\rightarrow}3$, baseline).
        Center: the same $M_3$ model evaluated with the unseen merge size~4
        ($3{\rightarrow}4$, length generalization).
        Right: $M_4$, trained and evaluated with maximum merge size~4
        ($4{\rightarrow}4$, native reference).
        \textcolor{baseblue}{\textbf{Blue}}: base tokens.
        \textcolor{mergeyellow}{\textbf{Yellow}}: hypertokens (composed of 2 base tokens).
        \textcolor{mergeorange}{\textbf{Orange}}: hypertokens (composed of 3 base tokens).
        \textcolor{mergered}{\textbf{Red}}: hypertokens (composed of 4 base tokens).
    }
    \label{fig:hypertoken-length-generalization-example}
\end{figure}

\paragraph{Evaluation V: Hypertoken Length Generalization.}
Because the hierarchical encoder applies one shared pairwise composer at every
recursion depth (\Cref{sec:merge-size-transfer}), it can compose hypertokens longer
than any it saw in training without adding parameters. We train with maximum merge
size 3 and raise the cap to 4 at inference time.
\Cref{fig:hypertoken-length-generalization-example} shows that the model does form
and use the unseen length-4 merges, and quantitatively it tracks a model trained
natively at size 4 closely: on WikiText the byte-perplexity gap is under $0.1\%$,
and it stays within a few percent even on repeated-window text engineered to make
length-4 merges frequent. Extrapolating further, to merge sizes $5,\ldots,8$, the
largest jump is at the training boundary and performance plateaus after it. Full
numbers and the stress-test construction are in \Cref{app:merge_size_transfer}.

\section{Conclusion}

In this work we investigate the performance loss incurred when a pretrained language
model is upgraded from its static BPE vocabulary to an LZW-based dynamic one. We
find that a series of architectural and algorithmic design choices in prior work are
suboptimal and we propose an effective alternative for each.
The resulting recipe strongly outperforms the original \texttt{zip2zip} on every
perplexity corpus and at both model scales, and roughly halves the distance to an
uncompressed model. Although the gap is not yet closed, this is
a substantial step towards making adaptive tokenization quality-neutral.

\section*{Acknowledgments}
This work was supported as part of the Swiss AI Initiative by a grant from the Swiss National Supercomputing Centre (CSCS) under project ID \texttt{a0101} on Alps.
West's lab is partly supported by grants from the Swiss National Science Foundation (200021\_185043), the Swiss Data Science Center (P22\_08), H2020 (952215), the Microsoft Swiss Joint Research Center, and Google, and by generous gifts from Facebook, Google, and Microsoft.

\subsection*{AI use statement}

We used generative AI tools, both agentic assistants and code completion, for
(1)~programming, (2)~paper writing and mock reviewing, (3)~brainstorming and
scientific discussion, and (4)~literature review. We did not use them to run
experiments or to produce any reported number. We have reviewed all AI-assisted
work and take responsibility for the final content of this paper.

\subsection*{Ethics statement}

This work studies the token vocabulary of language models and introduces no new
data collection, human subjects or model capabilities beyond those of the
publicly released checkpoints it builds on. The datasets used are standard public
corpora and benchmarks. One consequence of the method is worth stating: a shorter
token sequence lowers inference cost, which makes the underlying models cheaper
to deploy for both beneficial and harmful uses alike.

\subsection*{Reproducibility statement}

The finetuning configuration, data mixture and evaluation details needed to
reproduce every number in this paper are given in \Cref{app:experiment_setup},
with per-benchmark settings in \Cref{app:benchmarks}, the GSM8K answer-extraction
rule in \Cref{app:answer_extraction}, and the fixed perplexity subsets in
\Cref{app:ppl_corpora}. \Cref{app:more_experiment_results} reports the full
per-component results, the seed spread and the branches that did not work. Our
Rust implementation of Fault-Tolerant LZW, the training code and the model
checkpoints are publicly available at \url{https://github.com/epfl-dlab/zip2zip-plus-plus}.

\bibliography{custom}
\bibliographystyle{iclr2027_conference}

\appendix
\section{Related Work}
\label{app:related_work}

\paragraph{Adapting a fixed vocabulary.}
The most direct response to over-segmentation is to change the vocabulary offline.
One line adds tokens for a target language or domain and continues pretraining:
Chinese~\citep{adapt-llama-chinese}, Korean~\citep{kim2024efficienteffectivevocabularyexpansion_korean},
and specialised domains such as mental health~\citep{expand-vocab-for-tasks,liu-etal-2024-gold},
with follow-up work studying which tokens are worth adding~\citep{language-adaptation-empirical-study,wang-etal-2019-improving}.
A second line improves the vocabulary-learning algorithm itself, relaxing BPE's
word-boundary constraint~\citep{gee-etal-2023-multi,liu-etal-2025-superbpe} or
choosing merges by an information criterion~\citep{goriely2025bytespaninformationdrivensubwordtokenisation},
which yields more compact tokenization than
BPE~\citep{sennrich-etal-2016-neural,kudo-richardson-2018-sentencepiece,kumar-thawani-2022-bpe}.
Both lines produce a vocabulary that is fixed once training ends, so they require
per-domain intervention and cannot exploit repetition inside a particular input.

\paragraph{Compressing the input.}
Prompt-compression methods shorten the context by discarding or summarising it:
gist tokens~\citep{NEURIPS2023_3d77c6dc}, selective
context~\citep{li-etal-2023-compressing}, LLMLingua~\citep{jiang-etal-2023-llmlingua},
summary vectors~\citep{chevalier-etal-2023-adapting}, in-context
autoencoders~\citep{ge2024incontext} and soft-prompt
compression~\citep{wingate-etal-2022-prompt}. These achieve high compression
ratios precisely because they are \emph{lossy}, and they compress only the input:
generated tokens---which dominate latency at low batch size---are unaffected.
Closest in spirit to compressed language modelling is
\citet{lester2024trainingllmsneurallycompressed}, who train language models
directly on text compressed with arithmetic coding, and find that the resulting
codes are hard to model. LZW differs in that its codes stay aligned with the
original token boundaries, which is what allows an existing model to be adapted to
them rather than retrained from scratch.

\paragraph{Dynamic and byte-level architectures.}
A parallel line removes the fixed vocabulary from the architecture instead of the
tokenizer, learning where to segment as part of the model: Hourglass and dynamic
pooling~\citep{nawrot-etal-2022-hierarchical,nawrot-etal-2023-efficient},
MegaByte~\citep{yu2023megabyte}, SpaceByte~\citep{slagle2024spacebyte},
BLT~\citep{blt}, H-Net~\citep{hnet},
Toucan~\citep{fleshman2023toucantokenawarecharacterlevel},
Learn-Your-Token~\citep{thawani-etal-2023-learn},
ZeTT~\citep{NEURIPS2024_532ce4fc} and retrofitted dynamic
tokenization~\citep{feher-etal-2025-retrofitting}. These require training the
segmentation policy jointly with the model, and offer no lossless codec that maps
back to a canonical token stream. Dynamic Vocab~\citep{liu-etal-2024-generation}
shares our motivation most closely---it too grows the vocabulary during
generation---and differs in how entries are created and how the model is trained.
\texttt{zip2zip}~\citep{geng2025zip2zip} is the setting we build on: it keeps the
base BPE vocabulary, adds LZW-derived hypertokens on top of it, and adapts a
pretrained model with LoRA~\citep{hu2022lora}.

\paragraph{The interfaces a compressed stream disturbs.}
Our three architectural findings each touch a well-studied component. Tying the
input and output embedding matrices is standard for
small models~\citep{press-wolf-2017-using,inan-etal-2017-tying}; we show it fails
for a prefix-closed dynamic vocabulary, where the two matrices are pushed toward
incompatible geometries. Rotary position embeddings~\citep{su-etal-2024-roformer}
assume one position per token, which a variable-span hypertoken violates. And the
fragility of digit tokenization for arithmetic is
documented~\citep{nogueira-etal-2021-arithmetic,singh2024tokenizationcountsimpacttokenization};
we find it is the single largest source of degradation under compression, and that
exempting digits from merging recovers most of it.

\paragraph{Tokenization ambiguity and marginal likelihood.}
That a text has many valid tokenizations, and that scoring only the canonical one
understates a model's likelihood, is known for static vocabularies.
\citet{cao-rimell-2021-marginal} argue that language models should therefore be
evaluated on the marginal likelihood over tokenizations, while subword
regularization~\citep{kudo-2018-subword} and BPE-dropout~\citep{provilkov-etal-2020-bpe}
exploit the same multiplicity on the training side, sampling among segmentations for
robustness. For a fixed BPE vocabulary, however, the effect turns out to be small:
\citet{chirkova-etal-2023-marginalize} estimate the marginal probability by
importance sampling and find the log-likelihood gap is under $0.5\%$ in most cases.
Our setting is qualitatively different. A prefix-closed LZW dictionary is the
maximal violation of the prefix-free condition, so the number of valid segmentations
grows exponentially in the span length instead of comprising a few alternatives, and
the model is free to emit any of them. Our multi-view perplexity
(\Cref{sec:fault-tolerant}) marginalises over that set exactly, in one packed
forward pass rather than by sampling, and finds a gap of a wholly different
order---about a third of the apparent perplexity deficit of a compressed model (29--36\% of the strict deficit of \texttt{zip2zip++} against its uncompressed control, \Cref{app:multiview_scoring}) is a
scoring artefact rather than a modelling failure.

\paragraph{Other routes to fewer forward passes.}
Speculative decoding~\citep{leviathan2023fastinferencetransformersspeculative,chen2023acceleratinglargelanguagemodel}
also emits several tokens per target-model pass, but leaves the vocabulary and the
output distribution untouched: it trades extra compute for latency and is exact by
construction. Adaptive tokenization is complementary---it shortens the sequence
itself, which reduces KV-cache and context consumption as well as decoding steps,
at the cost of having to preserve model quality across a changed vocabulary.
\section{Embeddings Probes}
\label{app:embedding_probes}

\begin{figure}[h]
\centering
\includegraphics[width=\textwidth]{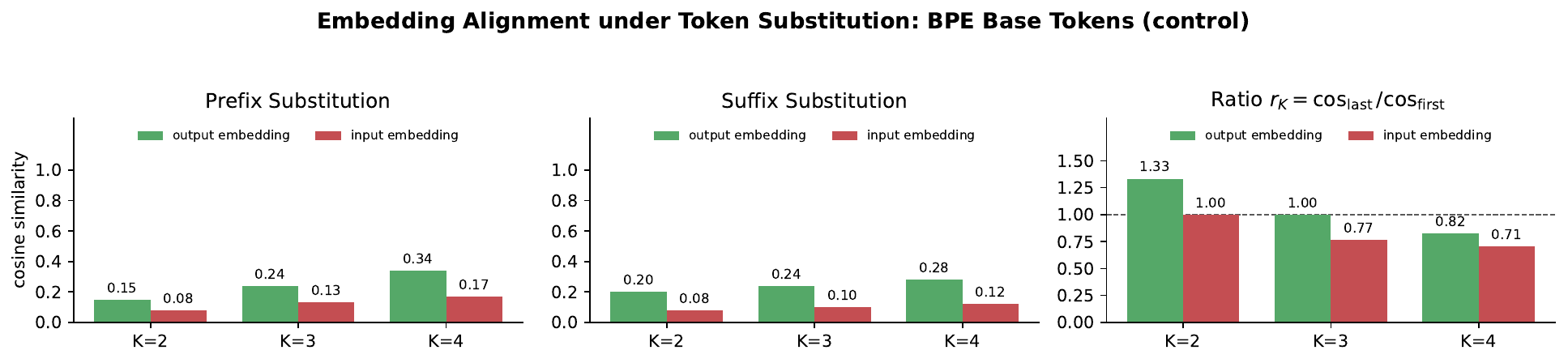}
\caption{\textbf{BPE base-table control for the substitution probe} (zip2zip++, Phi-3.5-mini).
The identical probe applied directly to the pretrained $\mathbf{W}_{\text{in}}$
and $\mathbf{W}_{\text{out}}$, with no hyper-encoder in the path. Panels and
metrics match \Cref{fig:substitution-barplot}. Both tables stay within
$r_K \in [0.71, 1.33]$ at every $K$, i.e.\ no asymmetry comparable to the
$4.8$--$6.6$ of the output hyper-encoder.}
\label{fig:substitution-barplot-bpe}
\end{figure}

\paragraph{Substitution Details.}
We sample the top 5{,}000 hypertokens (by corpus frequency) for each
span length $K \in \{2, 3, 4\}$ from WikiText-2, excluding spans containing
special or digit tokens. As a control, we run identical diagnostics directly on
the raw embedding matrices $\mathbf{W}_{\text{in}}$ and
$\mathbf{W}_{\text{out}}$ (no encoder) to verify that any observed structure is
created by the hyper-encoders rather than inherited from the pretrained
tables.
\Cref{fig:substitution-barplot-bpe} reports this control. Unlike the LZW
hypertoken embeddings of \Cref{fig:substitution-barplot}, the raw BPE tables are
close to positionally symmetric: $r_K$ stays within $0.71$--$1.33$ for both
matrices at every $K$, against $4.8$--$6.6$ for the output hyper-encoder. The
mild residual drift below~1 at larger $K$ reflects nested real words sharing
more suffix than prefix mass, not a compositional strategy. The prefix-aligned and balanced geometries
reported in \Cref{sec:untied} are therefore properties the
hyper-encoders acquire during training, not artifacts of the pretrained
embedding space.

\begin{figure*}[t]
    \centering
    \includegraphics[width=\textwidth]{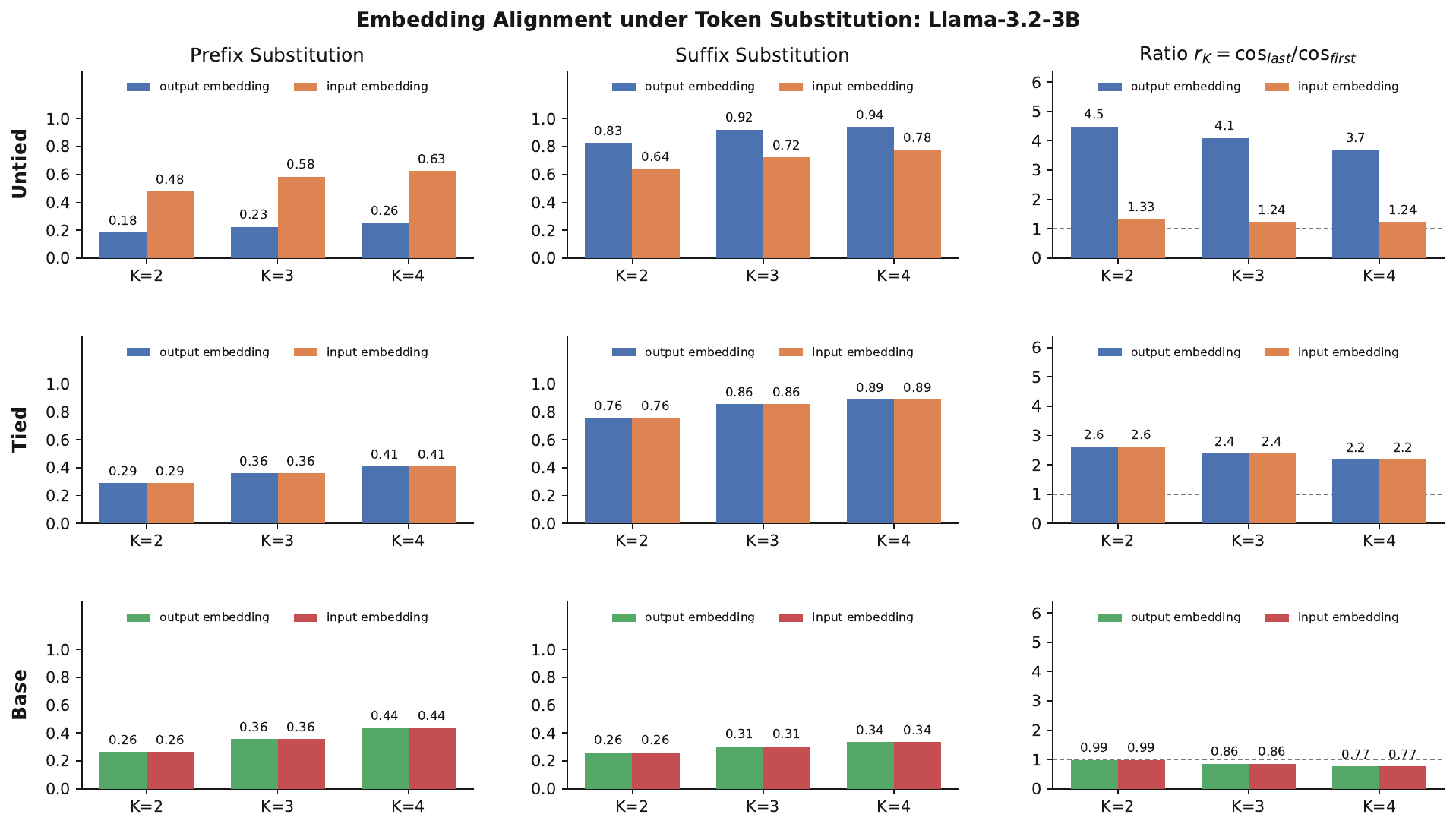}
    \caption{
        \textbf{Substitution probe on Llama-3.2-3B hypertokens.}
        The untied output hyper-encoder is strongly prefix-aligned, whereas
        the untied input hyper-encoder exhibits substantially weaker positional
        asymmetry. When the two hyper-encoders are tied, their input and output ratios coincide by construction and consistently fall between
        the corresponding untied output and input ratios.
        The base-token embeddings remain close to or below
        $r_K = 1$, showing no comparable prefix-aligned asymmetry.
    }
    \label{fig:llama-substitution-probe}
\end{figure*}

\begin{figure}[t]
\centering
\includegraphics[width=0.8\textwidth]{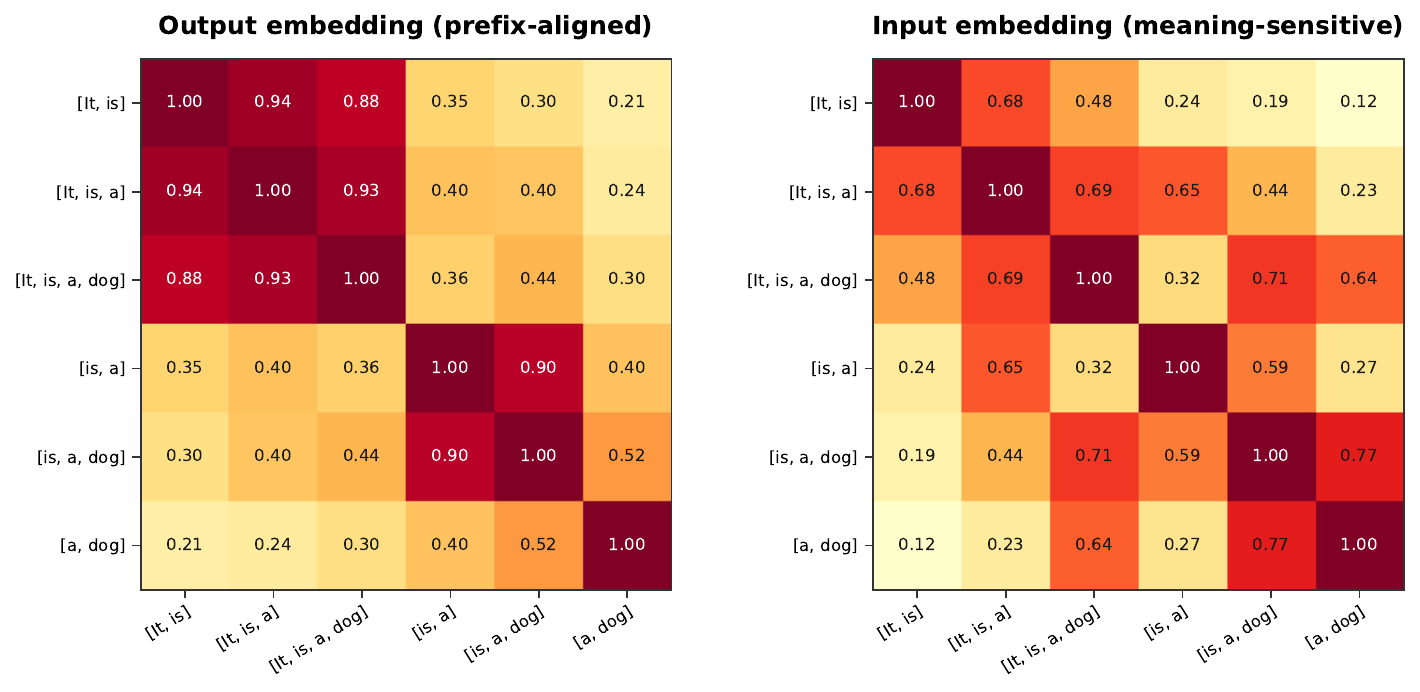}
\caption{\textbf{Sequence probe: pairwise similarity of contiguous sub-spans}
(zip2zip++, Phi-3.5-mini). Cosine similarity between the embeddings of all contiguous sub-spans of
length $\geq 2$ of \textit{``It is a dog''}. \emph{Left:} the output
hyper-encoder shows clear block structure---spans sharing a prefix cluster at
$\cos \geq 0.88$ while cross-block pairs fall to $0.21$--$0.44$. \emph{Right:}
the input hyper-encoder shows no such blocks but a smooth gradient in the number
of shared constituents, consistent with the balanced geometry of
\Cref{sec:untied}.}
\label{fig:sequence-heatmap}
\end{figure}

\paragraph{Sequence probe.}
We complement the aggregate substitution probe with an illustrative span-level
analysis. For the 4-token span \textit{``It is a dog''}---written $ABCD$
below---we embed every contiguous sub-span of length $\geq 2$ and compute all
pairwise cosine similarities (\Cref{fig:sequence-heatmap}). This adjudicates
whether similarity follows a shared prefix or broader constituent overlap: a
prefix-aligned encoder should cluster the nested chain
$AB \subset ABC \subset ABCD$ tightly regardless of what follows, whereas a
balanced encoder should instead track how many constituents two spans have in
common.

Both predictions hold. In the output embedding the prefix chain forms a block at
$\cos \geq 0.88$, while every pair spanning the two blocks falls to
$0.21$--$0.44$. In the input embedding there is no such block structure but a
smooth gradient: $ABCD$ sits closer to $BCD$ ($0.71$), with which it shares three
constituents, than to its own prefix $AB$ ($0.48$), with which it shares two.

\section{Experimental Setup}
\label{app:experiment_setup}

\Cref{tab:finetune-hparams} lists the configuration, shared by every run in this
paper and by the uncompressed control. The control sets the codebook size to zero,
so compression becomes the exact identity, the hyper-encoder path is not executed,
and the objective reduces to ordinary next-token prediction. It is
\emph{update-matched} but not base-token- or trainable-parameter-matched: at equal
decoder positions the control sees ${\approx}262$M base tokens against
${\approx}375$M for the compressed recipes, which additionally train two
hyper-encoders. We report compression as
$C = N_{\mathrm{base}} / N_{\mathrm{compressed}} \geq 1$.

\paragraph{Implementation.}
All three are implemented behind one interface in a single training and evaluation
stack, so they share the data loader, the LoRA setup and the evaluation harness;
any difference between them is attributable to the training representation and to
the window filter it implies (\Cref{app:variance_comparability}), not to the pipeline.

\begin{table}[h]
    \centering\small
    \caption{Finetuning configuration, shared across all runs and the control.}
    \label{tab:finetune-hparams}
    \begin{tabular}{@{}ll@{}}
        \toprule
        Base models & Phi-3.5-mini-instruct (3.8B), Phi-3-medium-4k-instruct (14B), \\
                    & Llama-3.2-1B-Instruct, Llama-3.2-3B-Instruct \\
        \midrule
        Adapter & LoRA rank 32, $\alpha = 32$, all attention and MLP projections; \\
                & decoder otherwise frozen \\
        Optimizer & AdamW, $\beta = (0.9, 0.999)$, weight decay 0.01 \\
        Learning rate & cosine $3\times10^{-4} \rightarrow 10^{-5}$, 800 warm-up steps \\
        Steps & 8{,}000, at $32{,}768$ decoder positions per step \\
        Window & 2{,}048 tokens \\
        Seed & 42 \\
        \midrule
        Max merge size $M$ & 4 \\
        Codebook & 4{,}096 entries; a 2{,}048-token window creates at most 2{,}048 \\
                Protected set & special and added tokens, plus every vocabulary piece that is a run of \\
                      & ASCII digits (10 pieces for Phi, 1{,}110 for Llama-3.2, whose pre-tokenizer \\
                      & splits numbers into groups of up to three digits) \\
        \midrule
        Data & ${\sim}1$B-token instruction mixture (multi-turn chat, competition \\
             & mathematics with worked solutions, web and code text), re-tokenized \\
             & per model: 997M tokens (Phi), 846M (Llama), eight shards. Chat \\
             & documents use the model's own template, loss on assistant turns only. \\
        \midrule
        Evaluation & two-shot through the chat template, lm-evaluation-harness 0.4.12, \\
                   & seed 1234, length-normalized accuracy where applicable; \\
                   & byte-level perplexity over disjoint windows of 1{,}024 base tokens, \\
                   & each compressed independently with a fresh codebook \\
        \midrule
        Hardware & 4$\times$H100 80GB; 1.5\,h (Llama-1B), 2.5\,h (Llama-3B, \\
                 & Phi-3.5-mini), 6.5\,h (Phi-3-medium), plus evaluation \\
                         \midrule
        Hyper-encoders & two flat pre-norm transformer layers over the ${\le}M$ constituent \\
                       & embeddings with a learned slot embedding; width = decoder width, \\
                       & 32 heads (40 for Phi-3-medium), $4\times$ feed-forward; non-causal, \\
                       & mean-pooled, added to the first constituent's embedding \\
        \bottomrule
    \end{tabular}
\end{table}

\subsection{Benchmarks and evaluation details}
\label{app:benchmarks}

\paragraph{Multiple-choice tasks.}
ARC-Challenge, ARC-Easy, HellaSwag, OpenBookQA, PIQA and WinoGrande are evaluated
two-shot through the chat template, with the few-shot examples supplied as previous
turns. We report length-normalised accuracy for all of them except WinoGrande,
where normalisation does not apply. Bootstrapped standard errors are at most
$0.022$ on OpenBookQA and below $0.015$ on the other five tasks.

\paragraph{Generation tasks.}
GSM8K is two-shot, capped at 256 emitted tokens (a hypertoken counts once, so a
compressed model may decode up to $C_{\text{gen}}$ times more base tokens within the
cap), with answers extracted as in \Cref{app:answer_extraction}.
MATH-500 uses the four-shot Minerva prompt and answers are checked symbolically
with \texttt{math\_verify}. HumanEval is zero-shot greedy \mbox{pass@1}, with
generation stopped at the closing code fence. IFEval is zero-shot prompt-level
strict accuracy.

\paragraph{Long context.}
RULER is evaluated at 4K, 8K, 16K and 32K context on a fixed subset of 200
examples per task. Because the post-SFT checkpoints were trained with 2{,}048-token
sequences and received no long-context training, RULER uses inference-time LongRoPE
adaptation with the official 128K configuration of the corresponding Phi model
family; no LongRoPE finetuning was performed.

\subsection{Answer extraction for GSM8K and MATH-500}
\label{app:answer_extraction}

Strict-format metrics require the answer template verbatim (GSM8K
\texttt{\#\#\#\#}, the Minerva \emph{Final Answer} sentence). Compressed models
rarely reproduce these templates even when their answers are correct: on
Phi-3.5-mini, GSM8K strict-match is .37/.04/.22 and MATH-500 exact-match
.12/.01/.01 for Uncompressed/\texttt{zip2zip}/\texttt{zip2zip++}, against .75/.36/.68
and .36/.15/.24 under flexible extraction. We therefore report flexible metrics
throughout and treat the strict gap as a formatting effect. Template adherence is
itself a facet of instruction following, and the IFEval drop of the compressed
models (\Cref{tab:post-sft}) points the same way; the flexible rule isolates
arithmetic correctness from this effect rather than denying it.

The flexible rule has a symmetric failure mode: the uncompressed 14B baseline
answers correctly and then, instead of stopping, continues with an unrelated new
problem in the style of its finetuning data. This happens in 275 of 1{,}319
generations and defeats a plain last-number heuristic, which scores that model .49
where the strict rule gives .66. Every GSM8K number in the main text and in \Cref{tab:strict_vs_flexible,tab:gsm8k_all,tab:finetune-scaling} therefore uses one rule: the answer is the number following ``\#\#\#\#'' when the model emits that
template, and the last number of the generation otherwise. Relative to plain
flexible extraction this moves the 14B baseline from .49 to .70 and no other row by
more than half a point. The evaluation-only diagnostics of \Cref{tab:selective-protection} use plain flexible extraction, as their generations were not stored; on the checkpoints where both rules are available they differ by at most 0.5 points.

\begin{table}[H]
\centering\small
\caption{GSM8K under the two extraction rules along the ablation ladder of
\Cref{tab:ablation}. The flexible column uses the combined rule defined above.
Strict matching demands the \texttt{\#\#\#\#} template verbatim; it moves by 14 points between the first two rungs, whose combined-rule scores differ by 2.7, which is why we report the combined rule throughout.}
\label{tab:strict_vs_flexible}
\begin{tabular}{@{}lcc@{}}
\toprule
Recipe & GSM8K flexible & GSM8K strict \\
\midrule
Protected digits, untied encoders & .6232 & .066 \\
\quad $+$ end-indexed positions   & .6497 & .210 \\
\quad $+$ identity init.          & .6778 & .215 \\
\midrule
Uncompressed reference            & .7468 & .372 \\
\bottomrule
\end{tabular}
\end{table}

\subsection{Perplexity corpora and the fixed subsets}
\label{app:ppl_corpora}

WikiText is the full WikiText-2 (raw) test set (62 documents, the
lm-evaluation-harness \texttt{wikitext} task); Pile is the validation split of The
Pile; mC4 and dC4 are the \texttt{en} and \texttt{en.noblocklist} validation splits
of C4, the latter without bad-word filtering. The Pile, mC4 and dC4 validation sets
are too large to score for every checkpoint at 1{,}024-token context, so all rows
use the same fixed subset of 1{,}000 documents per corpus, drawn once with seed
1234 and pinned by content hash; scores are reproducible to $10^{-4}$ (bf16 jitter
${\approx}10^{-5}$). On the four Phi-3.5-mini rows the subset values differ from
the full-corpus values by at most 0.04 byte-perplexity and never change the
ordering between rows.
\section{More Experimental Results}
\label{app:more_experiment_results}

Every table in this section uses the setup of \Cref{app:experiment_setup}.
Where a compressed-trained checkpoint is evaluated twice, once with compression
enabled and once disabled, we decompose the gap to the uncompressed control as
\begin{equation}
\label{eq:gap-decomposition}
S_{\mathrm{ctrl}}-S_{\mathrm{comp}}
=
\underbrace{S_{\mathrm{ctrl}}-S_{\mathrm{base}}}_{\text{weight-adaptation cost}}
+
\underbrace{S_{\mathrm{base}}-S_{\mathrm{comp}}}_{\text{representation cost}},
\end{equation}
where $S_{\mathrm{base}}$ is the score of the compressed-trained weights with
compression disabled. This separates damage stored in the adapted weights from
damage caused by presenting those weights with a compressed sequence.

\subsection{Main Results, All Models}
\label{app:main_results_full}

\Cref{tab:nlp_perplexity_full} extends \Cref{tab:nlp_perplexity} to all four base
models and adds the \textbf{Original} row: the released instruct checkpoint with its
own tokenizer, evaluated without any finetuning. Comparing \textbf{Original} with
\textbf{Uncompressed} isolates what our finetuning mixture does on its own, before
compression enters: on the multiple-choice tasks it helps Phi-3.5-mini and Llama-3.2-1B slightly, is roughly
neutral on Phi-3-medium, and costs Llama-3.2-3B a few points on ARC-Challenge; on GSM8K it costs every model 3--14 points
(\Cref{tab:gsm8k_all}), most on Phi-3-medium. The compressed rows should therefore be read against \textbf{Uncompressed} rather than
against the released model.

\begin{table*}[htbp]
    \centering
    \small
    \setlength{\tabcolsep}{3.5pt}
    \caption{\small \textbf{Main results, all four base models.} Columns and
    conventions as in \Cref{tab:nlp_perplexity}. \textbf{Original} is the released
    instruct checkpoint with no finetuning; \textbf{Uncompressed} is the same model
    finetuned on uncompressed sequences. Bold marks the better of the two
    compressed models in each column.}
    \label{tab:nlp_perplexity_full}
    \begin{tabular}{ll|cccccc|cccc}
        \toprule
        \textbf{Model} & \textbf{Method}
        & {ARC-c} & {ARC-e} & {HS} & {OBQA} & {PIQA} & {WG}
         & {Wiki} & {Pile} & {mC4} & {dC4} \\
        \midrule
        \multirow{4}{*}{Phi-3.5-4B}
        & Original  & 0.56 & 0.79 & 0.71 & 0.48 & 0.78 & 0.73 & 1.58 & 1.68 & 1.79 & 1.84 \\
        & Uncompressed  & 0.59 & 0.84 & 0.74 & 0.48 & 0.82 & 0.75 & 1.58 & 1.67 & 1.77 & 1.82 \\
        & zip2zip   & 0.54 & 0.81 & 0.71 & 0.46 & \textbf{0.80} & \textbf{0.75} & 1.70 & 1.82 & 1.86 & 1.92 \\
        & zip2zip++ & \textbf{0.57} & \textbf{0.83} & \textbf{0.72} & \textbf{0.47} & \textbf{0.80} & 0.74 & \textbf{1.66} & \textbf{1.78} & \textbf{1.83} & \textbf{1.89} \\
        \hline
        \multirow{4}{*}{Phi-3-14B}
        & Original  & 0.64 & 0.86 & 0.82 & 0.57 & 0.83 & 0.79 & 1.43 & 1.62 & 1.72 & 1.77 \\
        & Uncompressed  & 0.65 & 0.88 & 0.81 & 0.51 & 0.83 & 0.79 & 1.43 & 1.61 & 1.71 & 1.76 \\
        & zip2zip   & 0.61 & 0.85 & 0.78 & \textbf{0.51} & 0.81 & 0.77 & 1.58 & 1.76 & 1.80 & 1.85 \\
        & zip2zip++ & \textbf{0.62} & \textbf{0.86} & \textbf{0.79} & 0.50 & \textbf{0.82} & \textbf{0.78} & \textbf{1.53} & \textbf{1.72} & \textbf{1.77} & \textbf{1.82} \\
        \hline
        \multirow{4}{*}{Llama-3.2-1B-Inst}
        & Original  & 0.39 & 0.63 & 0.58 & 0.35 & 0.74 & 0.58 & 1.74 & 1.79 & 1.88 & 1.94 \\
        & Uncompressed  & 0.39 & 0.66 & 0.59 & 0.35 & 0.75 & 0.60 & 1.71 & 1.77 & 1.84 & 1.89 \\
        & zip2zip   & \textbf{0.38} & \textbf{0.64} & \textbf{0.58} & \textbf{0.37} & \textbf{0.74} & \textbf{0.61} & 1.78 & 1.87 & 1.90 & 1.96 \\
        & zip2zip++ & 0.36 & \textbf{0.64} & \textbf{0.58} & 0.37 & \textbf{0.74} & \textbf{0.61} & \textbf{1.77} & \textbf{1.85} & \textbf{1.89} & \textbf{1.95} \\
        \hline
        \multirow{4}{*}{Llama-3.2-3B-Inst}
        & Original  & 0.51 & 0.77 & 0.69 & 0.43 & 0.77 & 0.65 & 1.65 & 1.71 & 1.78 & 1.84 \\
        & Uncompressed  & 0.47 & 0.77 & 0.69 & 0.43 & 0.78 & 0.68 & 1.63 & 1.68 & 1.74 & 1.80 \\
        & zip2zip   & \textbf{0.48} & \textbf{0.76} & 0.67 & \textbf{0.43} & \textbf{0.76} & \textbf{0.68} & 1.72 & 1.81 & 1.83 & 1.89 \\
        & zip2zip++ & \textbf{0.48} & \textbf{0.76} & \textbf{0.68} & \textbf{0.43} & \textbf{0.76} & \textbf{0.68} & \textbf{1.68} & \textbf{1.75} & \textbf{1.79} & \textbf{1.84} \\
        \bottomrule
    \end{tabular}
\end{table*}

\begin{table}[H]
    \centering\small
    \caption{GSM8K accuracy (two-shot, combined extraction rule of
    \Cref{app:answer_extraction}) and generation compression $C_{\text{gen}}$
    (base tokens per emitted token over the generated answers) for all four
    models. \textbf{Original} uses flexible extraction; the two rules give
    identical scores on the released Llama models. Bold marks the better of the
    two compressed models.}
    \label{tab:gsm8k_all}
    \begin{tabular}{@{}lcccc|cc@{}}
        \toprule
        & \multicolumn{4}{c|}{GSM8K accuracy$\uparrow$} & \multicolumn{2}{c}{$C_{\text{gen}}\uparrow$} \\
        & Original & Uncompressed & zip2zip & zip2zip++ & zip2zip & zip2zip++ \\
        \midrule
        Phi-3.5-mini      & .78 & .75 & .36 & \textbf{.68} & 1.53 & 1.25 \\
        Phi-3-medium      & .84 & .70 & .51 & \textbf{.69} & 1.52 & 1.27 \\
        Llama-3.2-1B-Inst & .42 & .34 & .20 & \textbf{.24} & 1.37 & 1.26 \\
        Llama-3.2-3B-Inst & .76 & .64 & .41 & \textbf{.53} & 1.37 & 1.29 \\
        \bottomrule
    \end{tabular}
\end{table}

\subsection{Protected Tokens}
\label{app:protected_tokens}

FT-LZW's protected-token set doubles as a diagnostic: by exempting one class of
tokens at a time we can ask \emph{which} tokens compression actually harms.
\Cref{tab:selective-protection} does this for digits, operators and \LaTeX{}
control tokens. Digits dominate, and adding the other two classes buys nothing
further, so effective protection has to be targeted rather than broad:
compressibility is not semantic substitutability.

\paragraph{Achieved compression.}
The compression \texttt{zip2zip++} actually realises is uneven across workloads: it
represents WikiText in $14\%$ fewer tokens ($1.17\times$) and its own GSM8K
generations in $20\%$ fewer ($1.25\times$), while the short two-shot
multiple-choice prompts shrink by only $7\%$---these tasks are both where
compression helps least and where the accuracy differences between recipes are
smallest. Part of the gap to the unprotected recipe, which compresses WikiText by
$1.21\times$, is the price of protecting digits, quantified in the last three
columns of \Cref{tab:selective-protection}. The same pattern holds on GSM8K
generations for every model (\Cref{tab:gsm8k_all}, last two columns):
\texttt{zip2zip} compresses answers by $1.37$--$1.53\times$ and \texttt{zip2zip++}
by $1.25$--$1.29\times$; the Llama models, whose BPE vocabulary already merges up to
three digits, gain the least from LZW on arithmetic text.

\begin{table}[H]
\centering\small
\setlength{\tabcolsep}{5pt}
\caption{\textbf{Token protection with FT-LZW.} Rows sharing a checkpoint differ
only in the evaluation representation, so the block structure instantiates
\Cref{eq:gap-decomposition}: protecting digits at evaluation alone recovers 10.2 of
the 12.7-point representation penalty, and protecting them during training as well
repairs both terms, a 25.4-point gain over the unprotected baseline. All compressed rows in this paper use the FT-LZW codec and differ only in its protected-token set. The cost is
localized to math-heavy generation; general-text and multiple-choice compression are
barely affected. Protecting operators and \LaTeX{} control tokens in addition yields
$+0.5$ points, confirming that protection should target empirically fragile atoms
rather than be applied broadly.}
\label{tab:selective-protection}
\label{tab:finetune-decomposition-full}
\begin{tabular}{@{}l c ccc@{}}
\toprule
& GSM8K$\uparrow$ & \multicolumn{3}{c}{Compression $C\uparrow$} \\
\cmidrule(lr){2-2}\cmidrule(lr){3-5}
Protection strategy & flexible & WikiText & MC input & Math decode \\
\midrule
No protection                                    & .356 & $1.209$ & $1.079$ & $1.53$ \\
\quad same weights, digits protected at eval     & .458 & $1.169$ & $1.077$ & $1.30$ \\
\quad same weights, uncompressed eval            & .483 & $1.000$ & $1.000$ & $1.00$ \\
\midrule
Digits protected (train $+$ eval)                & .610 & $1.169$ & $1.077$ & $1.28$ \\
\quad same weights, uncompressed eval            & .679 & $1.000$ & $1.000$ & $1.00$ \\
\quad $+$ operators and \LaTeX{} (eval only)     & .615 & $1.166$ & $1.077$ & $1.23$ \\
\midrule
Uncompressed control                             & .742 & $1.000$ & $1.000$ & $1.00$ \\
\bottomrule
\end{tabular}
\end{table}

\subsection{Per-Component Gains and Significance}
\label{app:ablation_stats}

Along the ladder of \Cref{tab:ablation}, digit protection accounts for most of the
GSM8K recovery ($+25.4$ points), and end-indexed positions and the
zero-initialized residual for most of the rest ($+2.7$ and $+2.8$ points). Untying the
hyper-encoders acts mainly on the other metrics ($+1.6$ points of mean
multiple-choice accuracy, $-0.025$ WikiText byte-perplexity, on unrounded scores).
Under an exact paired McNemar test on the 1{,}319 GSM8K problems,
\texttt{zip2zip++} is $32.2$ points above the unmodified \texttt{zip2zip} recipe
($p = 1\times10^{-83}$) and $6.9$ points below the update-matched
\texttt{Uncompressed} reference ($p = 3\times10^{-7}$): both the improvement and the
residual gap are real rather than noise. A second training seed of the final recipe
lands $1.8$ points below the first ($p = 0.14$), the same order as every individual
step after digit protection. We therefore read the ladder as a cumulative effect
that is significant end to end, rather than as a sequence of individually
significant steps. On the Llama models the same test gives \texttt{zip2zip++} vs.\ \texttt{zip2zip}
$+4.7$ points ($p{=}1.5{\times}10^{-4}$) at 1B and $+12.0$
($p{=}1.3{\times}10^{-17}$) at 3B, and $-9.3$ / $-11.6$ points vs.\
\texttt{Uncompressed} ($p{<}10^{-11}$).

\subsection{Multi-View Scoring}
\label{app:multiview_scoring}

\Cref{sec:fault-tolerant} defines multi-view perplexity as the exact sum over all
complete segmentations. Some of our earlier evaluations instead used a
\emph{first-token bound}, which credits the base token $x_1$ plus every available
hypertoken whose expansion is a nonempty prefix of the remaining target:
\begin{equation}
\label{eq:multiview-score}
    p^{*}_{\mathrm{ft}}(x_{1:n}\mid \mathbf{z}_{<i})
    = p(x_1\mid \mathbf{z}_{<i})
    + \!\!\sum_{z:\,\operatorname{expand}(z)\preceq x_{1:n}}\!p(z\mid \mathbf{z}_{<i}).
\end{equation}
Since the strict target is included,
$p^{*}_{\mathrm{ft}} \ge p_{\text{multi-view}} \ge p_{\mathrm{strict}}$. We report
both because the difference is large: exact marginalisation removes 29--36\% of the
canonical penalty, whereas the bound appears to remove 74--97\%. Segmentation
ambiguity is therefore a real but minority component of the perplexity deficit, and
we no longer use the bound as evidence.

\begin{table}[H]
    \centering\small
    \caption{Byte-perplexity of \texttt{zip2zip++} (Phi-3.5-mini) against the
    update-matched control, on WikiText and the fixed 1{,}000-document Pile/mC4/dC4
    subsets of \Cref{app:ppl_corpora}.}
    \label{tab:finetune-multiview}
    \label{tab:finetune-multiview-full}
    \begin{tabular}{@{}lrrrr@{}}
        \toprule
        Corpus & Control & Strict & Exact multi-view & First-token bound \\
        \midrule
        WikiText & 1.5777 & 1.6573 & 1.6290 & 1.5882 \\
        Pile     & 1.6678 & 1.7819 & 1.7488 & 1.6972 \\
        mC4      & 1.7660 & 1.8311 & 1.8084 & 1.7696 \\
        dC4      & 1.8221 & 1.8913 & 1.8668 & 1.8240 \\
        \bottomrule
    \end{tabular}
\end{table}

\subsection{Merge-Size Transfer}
\label{app:merge_size_transfer}

\begin{figure*}[t]
    \centering
    \includegraphics[width=\textwidth]
    {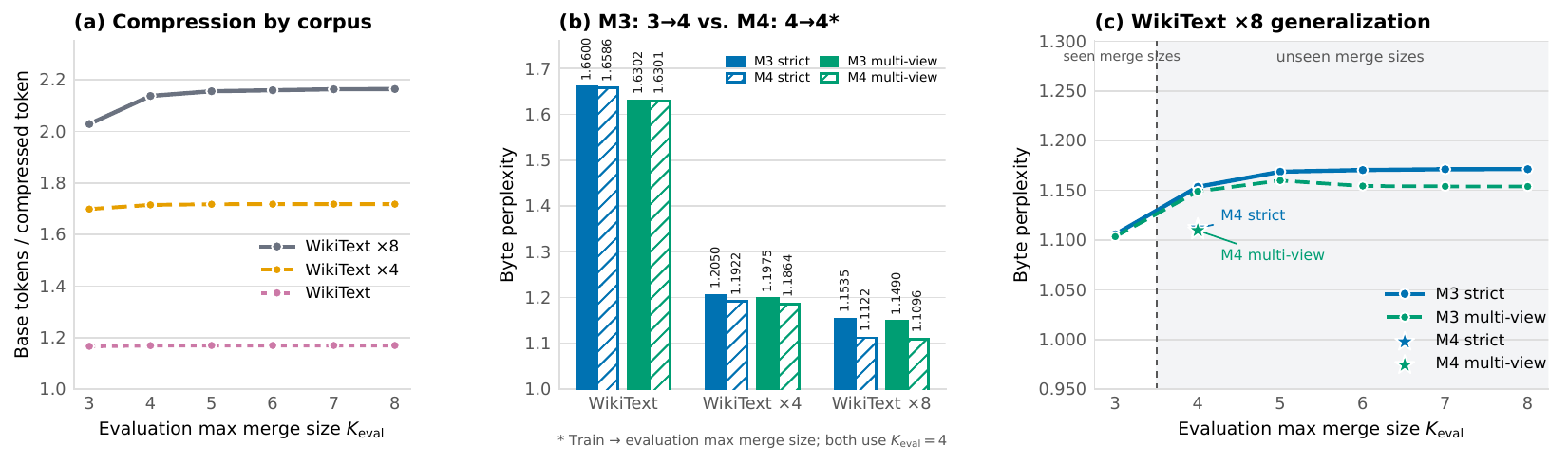}
    \caption{
    \textbf{Hierarchical merge-size generalization.}
    \textbf{(a)} Compression as the evaluation-time maximum merge size increases; WikiText ${\times}4$ and ${\times}8$ provide progressively stronger transfer exposure.
    \textbf{(b)} Strict and exact multi-view byte perplexity for transferred \(M_3:3{\rightarrow}4\) and native \(M_4:4{\rightarrow}4\).
    \textbf{(c)} \(M_3\), trained with \(K=3\), evaluated at \(K_{\mathrm{eval}}=3,\ldots,8\) on WikiText ${\times}8$, with native \(M_4\) at \(K_{\mathrm{eval}}=4\) for reference.
    }
    \label{fig:merge-size-transfer}
\end{figure*}

Natural text rarely exercises a larger merge-size cap: raising it from 3 to 4
changes compression by 0.018\% on our default evaluation and 0.22\% on WikiText.
We therefore stress-test on repeated-window WikiText, where length-4 spans cover
6.54\% or 29.90\% of scored base tokens; absolute perplexities on these synthetic
corpora are not comparable to natural WikiText. Under exact multi-view scoring the
transfer penalty is only 12\% and 5\% smaller than the strict penalty, so the
cross-size cost is a genuine limitation of the size-3 encoder rather than a
canonical-segmentation artefact --- the first-token bound would have hidden most of
it.

\begin{table}[H]
    \centering\small
    \setlength{\tabcolsep}{5pt}
    \caption{Merge-size transfer. $a\!\rightarrow\!b$ denotes training with maximum
    merge size $a$ and evaluating with size $b$. Penalties are for
    $3\!\rightarrow\!4$ relative to the native $4\!\rightarrow\!4$ model. Single
    runs.}
    \label{tab:merge-size-matched}
    \begin{tabular}{@{}lrrrrrrr@{}}
        \toprule
        & Length-4 & \multicolumn{3}{c}{Strict byte-PPL} & \multicolumn{3}{c}{Penalty of $3\!\rightarrow\!4$} \\
        \cmidrule(lr){3-5}\cmidrule(lr){6-8}
        Corpus & coverage & $3\!\rightarrow\!3$ & $3\!\rightarrow\!4$ & $4\!\rightarrow\!4$ & Strict & Exact MV & First-token \\
        \midrule
        Repeat-4 & 6.54\%  & 1.1890 & 1.2050 & 1.1922 & 1.07\% & .94\%  & .25\% \\
        Repeat-8 & 29.90\% & 1.1057 & 1.1535 & 1.1122 & 3.71\% & 3.54\% & .33\% \\
        \bottomrule
    \end{tabular}
\end{table}

\subsection{Data Scale}
\label{app:data_scale}

A raw-text pretraining stage before SFT improves language modeling but degrades
instruction-tuned reasoning, which longer SFT then partially restores while
surrendering most of the perplexity advantage (\Cref{tab:finetune-scaling}).
Extending SFT alone from 8k to 32k steps, for the digit-protected untied model
(third rung of \Cref{tab:ablation}, before end-indexed positions and
zero-initialisation), raises GSM8K from .623 to .672 and improves
WikiText byte-perplexity from 1.686 to 1.656---roughly what the full recipe reaches
at 8k steps with a quarter of the compute--- but the control was not extended, so
this is not an equal-compute result. The pretraining mixture, used at 20B and 200B
tokens (50\% FineWeb-Edu, 20\% Stack,
10\% FineMath, 20\% FineWeb2-HQ), compresses by only $1.3100\times$ despite its code
component, while the templated SFT mixture reaches $1.4504\times$: repeated dialogue
templates and mathematical markup matter more than semantic domain for online LZW
compressibility.

\begin{table}[H]
    \centering\small
    \caption{A raw-text pretraining stage before the standard SFT mixture. Single
    runs, without a matched long-run uncompressed control. The 200B row is the
    ``Continued Pretraining'' row of \Cref{tab:ablation}.}
    \label{tab:finetune-scaling}
    \begin{tabular}{@{}lrr@{}}
        \toprule
        Training path & GSM8K & Wiki byte-PPL \\
        \midrule
        zip2zip++, direct SFT, 8k steps & .6778 & 1.6574 \\
        20B raw $+$ SFT, 8k  & .6088 & 1.6377 \\
        20B raw $+$ SFT, 16k & .6255 & 1.6390 \\
        20B raw $+$ SFT, 48k & .6391 & 1.6554 \\
        200B raw $+$ SFT, 8k & .6414 & 1.6331 \\
        \bottomrule
    \end{tabular}
\end{table}

\subsection{Seed Variance and Comparability}
\label{app:variance_comparability}

\paragraph{Seed variance.}
Two runs of the final recipe give .6778 and .6596 GSM8K accuracy (combined rule), disagreeing on 240 of 1319 items (132 vs.\ 108, $p{=}0.14$), while their multiple-choice averages differ by .12
points and their final losses by $10^{-4}$. Strict accuracy moves in the opposite
direction ($+4.09$ points for the seed with lower flexible accuracy), confirming
that strict matching entangles reasoning with output format. Single-seed GSM8K
differences below roughly two points should therefore not be read as effects.

\paragraph{Comparability.}
Several factors limit which numbers may be compared directly. The control is
update-matched rather than base-token- or parameter-matched. Every run uses the same
data loader, which discards the ${\approx}6.5\%$ of training windows that compress by
more than $2\times$; the uncompressed control, whose ratio is 1, keeps every window.
Training the final recipe on the complete stream instead lowers GSM8K by 8.2 points
(.677${\to}$.595, paired McNemar $p{=}10^{-8}$) and worsens perplexity and generation
compression, so the filter is kept and reported here as a data choice rather than a
bug fix. A control trained on the same retained windows was not run, so the filter is
a confound shared by every compressed-versus-uncompressed comparison; and because the
retained set depends on the codec configuration (protected digits lower the ratio),
the rungs of \Cref{tab:ablation} need not train on identical windows. The released Llama models are scored through their own chat template, which
inserts the evaluation date into the system prompt; re-running GSM8K on a later date
moved their scores by 1.3--1.7 points, whereas our finetuned models pin the date and
re-run to within 0.9 points. We report achieved compression rather than wall-clock
decoding speed; the hyper-encoders and the codec add per-step costs that we have not
benchmarked. Finally, perplexity on Pile, mC4 and dC4 should be compared only within
a fixed sampling pipeline.

\subsection{Negative Results}
\label{app:negative_results}

\paragraph{What the recipe does not fix.}
The gains are concentrated in perplexity and mathematical reasoning. On HumanEval
and IFEval, \texttt{zip2zip++} is statistically indistinguishable from the
unmodified \texttt{zip2zip} recipe at both scales (paired McNemar $p = 0.16$ and
$p = 1$ on Phi-3.5-mini, $p = 0.32$ and $p = 0.16$ on Phi-3-medium): compression
costs 25--26 points of code generation and 6--9 points of instruction following, and
no statistically established part of it is recovered.

\paragraph{Branches that did not help.}
\Cref{tab:finetune-negative} lists the one-variable branches we tried. Several lower
the training loss or the canonical perplexity while lowering GSM8K, which is why we
report reasoning accuracy, decoded-text probability and achieved compression as
separate axes rather than relying on perplexity alone. Two rows need a caveat. By
default, compressed position $t$ may predict the base vocabulary and the codebook
entries with creation index $k \le t$ in its window; since LZW creates at most one
entry per consumed token, this is a superset of the entries that exist at $t$, so
some later-created entries enter the softmax denominator early, while the target
itself always exists. The same mask is used for training and for every reported
perplexity, strict and multi-view alike; generation masks to the entries present in
the decoder's codebook, which is exact by construction. Because the default set is a
superset of the entries the decoder can emit, restricting the softmax to the exact set
can only raise the target probability: for the same logits, the reported perplexities
of the compressed models are upper bounds on those of the deployed decoder. The online-mask run instead
exposes at each position exactly the entries created earlier in its window. It
improves every training-side proxy (teacher-forced next-token accuracy
$49.41 \rightarrow 50.15\%$, hypertoken accuracy $.420 \rightarrow .504$, generation
compression $1.2536 \rightarrow 1.3107\times$) yet worsens six of seven downstream
tasks. The two positional branches (two-axis RoPE puts even rotary pairs on
base-stream positions and odd pairs on compressed ones; the gated variant learns the
mix per layer) were still improving when the learning-rate budget ran out, so they
show that neither variant helps at 8k steps, not that compressed coordinates are
intrinsically harmful.

\begin{table}[H]
    \centering\scriptsize
    \caption{One-variable branches that did not improve the recipe. $\Delta$GSM is in
    percentage points relative to each branch's parent (the \texttt{zip2zip} baseline, the
    ``+ untied'' rung of \Cref{tab:ablation}, or the final recipe). ``Flat'' means no practically
    meaningful change. All are single runs; see \Cref{app:variance_comparability} for
    seed spread.}
    \label{tab:finetune-negative}
    \resizebox{\textwidth}{!}{%
    \begin{tabular}{@{}lllll@{}}
        \toprule
        Intervention & Parent & Training / LM proxy & $\Delta$GSM & Takeaway \\
        \midrule
                Reformat math SFT as chat            & \texttt{zip2zip} & Wiki PPL $1.695\!\rightarrow\!1.719$ & $+2.3$ & Narrow style gain, broad MC/PPL trade-off \\
        Freeze decoder LoRA for 200 steps    & + untied & Final loss flat & $-6.6$ & Co-adaptation is preferable here \\
        Hyper-encoder 2$\!\rightarrow\!$4 layers & + untied & Loss $1.51\!\rightarrow\!1.45$; PPL flat & $-2.6$ & Capacity fits training, does not transfer \\
        Exact online codebook mask           & final & Loss $1.434\!\rightarrow\!1.376$ & $-2.4$ & Better proxies, worse downstream \\
        25\% uncompressed microsteps         & final & PPL flat & $-3.6$ & No measurable transfer \\
        Two-axis RoPE (fixed 50/50 split)    & final & Wiki PPL flat & $-2.4$ & Mixed coordinates do not help \\
        Gated RoPE (512 low-freq.\ gates)    & final & \textbf{Best} Wiki PPL $1.6565$ & $-3.3$ & Best perplexity, worst reasoning \\
        Tie the two hyper-encoders           & final & MC/PPL near flat & $-9.1$ & Role-specific composition matters most \\
        Remove the residual connection       & final & Wiki PPL $1.657\!\rightarrow\!1.705$ & $-3.9$ & The pretrained anchor is important \\
        Hierarchical instead of flat encoder & final & Wiki PPL flat; slower & $-4.6$ & No gain at fixed $M$; kept for length transfer (\Cref{app:merge_size_transfer}) \\
        \bottomrule
    \end{tabular}}
\end{table}
\section{Fault-Tolerant LZW: Details}\label{app:ftlzw}

\paragraph{Where standard LZW breaks.}

\Cref{alg:lzw-contrast} isolates the single line that differs between the two
decoders. Standard LZW (red) inserts \texttt{buffer + z} unconditionally, because
a deterministic encoder guarantees that this combination is always new: every code
it emits creates exactly one dictionary entry. A neural compressor gives no such
guarantee---when it under-merges, \texttt{buffer + z} may already be present, and
the unconditional insert then writes a duplicate, desynchronising the decoder's
dictionary from the encoder's for the rest of the sequence.
FT-LZW (blue) replaces that line with a \emph{guarded merge}: it walks along
$\omega$, absorbing symbols whose combination with the buffer is already in the
dictionary, and inserts only when it reaches a genuinely novel one. On perfectly
compressed input the loop exits on its first iteration and the two decoders agree
exactly; on imperfectly compressed input FT-LZW skips the redundant insert
instead of corrupting the dictionary. Three further mechanisms---a disabled-token set, unknown-code
resolution and a codebook cap---are listed below.

\begin{algorithm}[H]
\caption{LZW decompression: standard (\textcolor{red}{red}) vs.\ FT-LZW
(\textcolor{blue}{blue}).}
\label{alg:lzw-contrast}
\begin{algorithmic}[1]
\State Initialize dictionary with the base tokens
       (e.g.\ \texttt{to}, \texttt{be}, \texttt{or}, \texttt{not})
       $\rightarrow$ \texttt{dict[0..3]}
\State \texttt{next\_id} $\leftarrow 4$
\State Read first code $X$;\quad \textbf{emit} \texttt{dict[}$X$\texttt{]};\quad
       \texttt{buffer} $\leftarrow$ \texttt{dict[}$X$\texttt{]}
\While{more codes}
    \State Read next code $Z$
    \State $\omega \leftarrow$ \texttt{dict[}$Z$\texttt{]};\quad
           \textbf{emit} $\omega$
    \State
    \textcolor{red}{%
    \State \textcolor{red}{$z \leftarrow$ first symbol of $\omega$}
        \Comment{\textcolor{red}{standard LZW: one unconditional insert}}
    \State \textcolor{red}{\texttt{dict[next\_id]} $\leftarrow$ \texttt{buffer} $+\, z$};\quad
           \textcolor{red}{\texttt{next\_id} $\leftarrow$ \texttt{next\_id} $+\, 1$}
    \State \textcolor{red}{\texttt{buffer} $\leftarrow \omega$}
    }%
    \State
    \textcolor{blue}{%
    \While{\textcolor{blue}{$\omega \neq \varepsilon$}}
        \Comment{\textcolor{blue}{FT-LZW: guarded merge}}
        \State \textcolor{blue}{$z \leftarrow$ first symbol of $\omega$}
        \If{\textcolor{blue}{\texttt{buffer} $+\, z \notin$ \texttt{dict}}}
            \State \textcolor{blue}{\texttt{dict[next\_id]} $\leftarrow$ \texttt{buffer} $+\, z$};\quad
                   \textcolor{blue}{\texttt{next\_id} $\leftarrow$ \texttt{next\_id} $+\, 1$}
            \State \textcolor{blue}{\texttt{buffer} $\leftarrow \omega$};\quad
                   \textbf{\textcolor{blue}{break}}
                \Comment{\textcolor{blue}{novel entry}}
        \EndIf
        \State \textcolor{blue}{\texttt{buffer} $\leftarrow$ \texttt{buffer} $+\, z$};\quad
               \textcolor{blue}{$\omega \leftarrow$ $\omega$ without its first symbol}
            \Comment{\textcolor{blue}{already known: absorb}}
    \EndWhile
    }%
\EndWhile
\end{algorithmic}
\end{algorithm}

\paragraph{A worked example.}

\begin{figure}[H]
    \centering
    \includegraphics[width=0.95\linewidth]{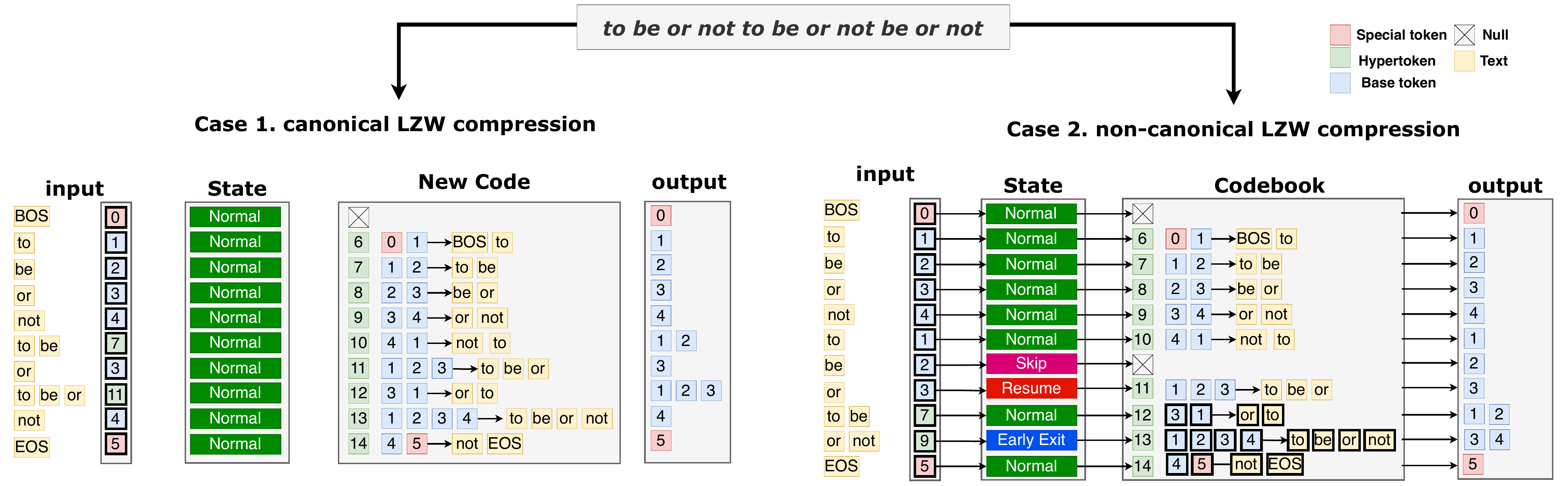}
    \caption{\textbf{FT-LZW decoding, step by step.} The same stream
    \emph{``to be or not to be or to be or not''} under canonical (left) and
    non-canonical (right) compression. Columns show the input codes, the decoder
    state, the codebook entry created at that step, and the emitted base tokens.
    Canonical input keeps the decoder in \textsc{Normal} at every step, one code
    creating one entry. Non-canonical input triggers the guard: \textsc{Skip}
    (no entry, the candidate already exists), \textsc{Resume} (the deferred entry
    is created on the next code), and \textsc{Early Exit} (a novel combination is
    reached partway through the expansion). Both cases emit the same base-token
    sequence.}
    \label{fig:ftlzw-trace}
\end{figure}

\Cref{fig:ftlzw-trace} traces the same token stream through both cases. Under
canonical compression every code creates exactly one new codebook entry, so the
decoder stays in its \textsc{Normal} state throughout. Under non-canonical
compression the guarded loop of \Cref{alg:lzw-contrast} takes its other branches:
it \textsc{Skip}s the insertion when the candidate entry already exists,
\textsc{Resume}s on the following code to create the entry that was deferred, and
takes an \textsc{Early Exit} when it reaches a novel combination partway through
$\omega$. In every case the emitted base-token stream is identical to the input,
which is what keeps decompression lossless even when the model under-merges.

\begin{itemize}[leftmargin=1.4em,itemsep=2pt,topsep=3pt]
    \item \textbf{Disabled tokens.} A designated set of token IDs (e.g.\
    \texttt{<pad>}, \texttt{<eos>}, and the protected tokens of \Cref{sec:lzw}) is
    emitted verbatim and resets the buffer, so these tokens are never absorbed into
    a composite entry.

    \item \textbf{Unknown-code resolution.} When a received code is not yet in the
    dictionary, two cases arise. If it equals the next free index, the decoder
    infers $\omega = W \!\cdot\! \mathrm{head}(W)$---the classic LZW ``cScSc''
    special case---and inserts and emits it immediately. If instead the buffer has
    reached the maximum entry length $M$, the unknown code must denote the buffer
    itself, force-emitted by the encoder on reaching the merge bound, so
    $\omega \gets W$.

    \item \textbf{Codebook cap.} Dictionary growth stops once $N$ new entries
    exist; the buffer resets and decoding continues without further growth,
    bounding the memory the codebook can consume over a long context.
\end{itemize}

\paragraph{Online variant for autoregressive generation.}

During generation the model emits one (hyper-) token at a time, so the dictionary
must be updated \emph{after each token} for newly created hypertokens to be
available at the next step. The online variant keeps a persistent state
$(W, D, j)$ across steps and runs exactly the guarded loop of
\Cref{alg:lzw-contrast} on each generated token; it differs from the batch decoder
only in that it receives one code per call and emits nothing, since the model has
already produced the token. Applying it token by token to a code sequence
therefore yields the same dictionary as decoding that sequence in one pass, which
is what keeps the encoder's and decoder's codebooks synchronized even when
sampling produces suboptimal compression.
\section{Tree Attention for Multi-View Scoring}
\label{app:tree_attention}

\begin{figure}[h]
    \centering
    \includegraphics[width=0.7\linewidth]{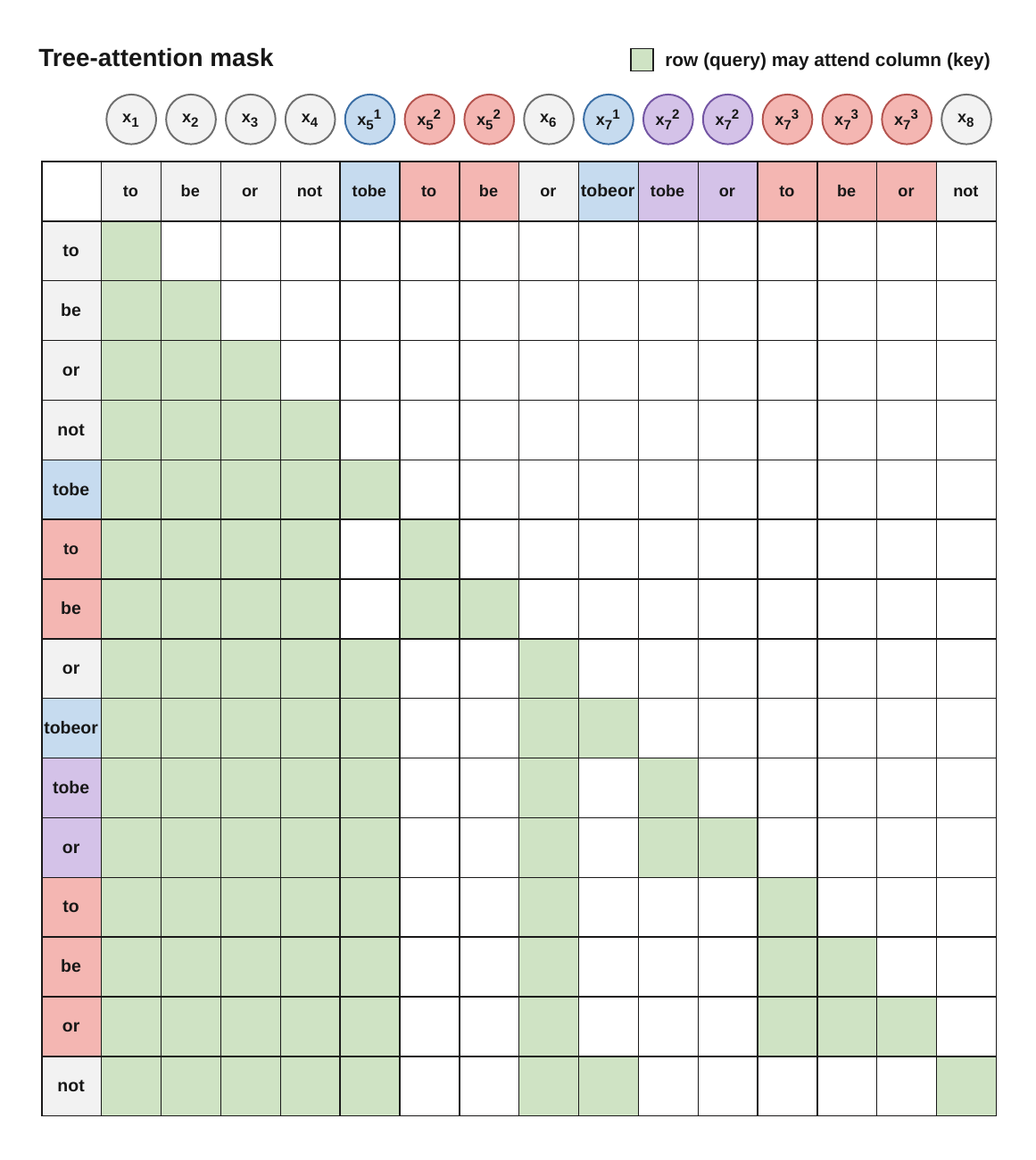}
    \caption{%
        \textbf{Tree attention for multi-view scoring.}
        The corresponding attention mask over the packed sequence.
        A shaded cell means the row (query) may read the column (key). The
        canonical backbone is causal; each branch sees only the shared prefix
        and its own ancestors, so sibling views and views of different targets
        cannot contaminate one another. All views are scored in one forward
        pass.%
    }
    \label{fig:tree-attention}
\end{figure}

The canonical sequence forms a backbone under ordinary causal attention. The distinct segmentation prefixes of each target are packed immediately after the corresponding root, forming a tree. The tree-attention mask allows each branch node to attend only to the canonical context preceding its target and
to its own ancestors; canonical nodes cannot see branches, and sibling or
cross-target branches remain isolated. Root probabilities are reused from the
canonical forward pass, while the packed pass computes all remaining
conditional probabilities. This matches independent prefix scoring up to numerical precision. On WikiText, tree attention is $26.0\times$ faster than the independent-prefix implementation and brings the total evaluation cost to only $1.6\times$ that of standard perplexity.

Every view is scored conditioned on the canonical prefix $z_{<i}$; alternative
segmentations of the prefix itself are not enumerated. The score is therefore
exact for the target given the canonical context and lies between the strict
score and the first-token bound (\Cref{app:multiview_scoring}), but it is not
the marginal over all segmentations of the whole text. The two coincide only if
the model's predictions do not depend on how the context was segmented; in
general the multi-view score can lie on either side of that marginal, so the
deficit that remains under it is a canonical-context measurement, not a bound
on the modelling cost of the compressed interface.

\section*{Author Contributions}
\begin{itemize}
    \item \textbf{Saibo Geng} led the project and the writing, designed the residual hyper-encoder, proposed multi-view perplexity and co-developed its tree-attention scoring, and co-developed FT-LZW and its open-source Rust library.
    \item \textbf{Andrea Gentilini} ran the main experiments, including continued pretraining, introduced digit protection and the protected-token analysis, developed size-aware RoPE, and co-led the evaluation.
    \item \textbf{Xinxian Ma} designed the hierarchical hyper-encoder and ran the merge-size generalization experiments, untied the hyper-encoders and analysed their embedding geometry, implemented multi-view perplexity together with its tree-attention scoring, and co-led the evaluation.
    \item \textbf{Nathan Ranchin} drove the early phase of the project, built the finetuning infrastructure, and co-developed FT-LZW and its open-source Rust library.
    \item \textbf{Yunzhen Yao} contributed to discussions throughout the project.
    \item \textbf{Robert West} supervised the project.
\end{itemize}
All authors contributed to writing the paper.

\end{document}